\documentclass[runningheads]{llncs}

\usepackage[mobile]{eccv}

\usepackage{eccvabbrv}

\usepackage{graphicx}
\usepackage{booktabs}

\usepackage[accsupp]{axessibility}

\usepackage{hyperref}

\usepackage{orcidlink}

\usepackage{makecell}
\usepackage[table]{xcolor}
\usepackage{multirow}
\usepackage{tabularx}
\usepackage[table]{xcolor}
\usepackage{booktabs}
\usepackage{algorithm}
\usepackage{algpseudocode}
\usepackage{bibunits}

\usepackage{graphicx}
\newsavebox\CBox
\def\textBF#1{\sbox\CBox{#1}\resizebox{\wd\CBox}{\ht\CBox}{\textbf{#1}}}
\newcolumntype{C}{>{\centering\arraybackslash}X}

\newcommand{\lightbottomrule}{\specialrule{\lightrulewidth}{\aboverulesep}{0pt}}
\newcommand{\zerotoprule}{\specialrule{0pt}{0pt}{\belowrulesep}}
\begin{document}

% ---------------------------------------------------------------
\title{Zero-Shot Depth from Defocus} 
\titlerunning{Zero-Shot Depth from Defocus}

\author{Yiming Zuo\textsuperscript{*} \and
Hongyu Wen\textsuperscript{*} \and
Venkat Subramanian\textsuperscript{*} \and Patrick Chen \and\\ Karhan Kayan \and Mario Bijelic \and Felix Heide \and Jia Deng}

\authorrunning{Y.Zuo, H.Wen, V.Subramanian et al.}
\institute{Department of Computer Science, Princeton University \\ \email{\{zuoym,hongyu.wen,venkat.subra,patrickchen,\\karhan,mario.bijelic,fheide,jiadeng\}@princeton.edu}}

\newcommand{\benchmarkname}{ZEDD}
\newcommand{\methodname}{FOSSA}

\maketitle
\let\thefootnote\relax\footnotetext{*Equal contribution.}

\vspace{-3mm}
\begin{abstract}
  Depth from Defocus (DfD) is the task of estimating a dense metric depth map from a focus stack. Unlike previous works overfitting to a certain dataset, this paper focuses on the challenging and practical setting of zero-shot generalization. We first propose a new real-world DfD benchmark \benchmarkname, which contains $8.3\times$ more scenes and significantly higher quality images and ground-truth depth maps compared to previous benchmarks. We also design a novel network architecture named \methodname. \methodname\ is a Transformer-based architecture with novel designs tailored to the DfD task. The key contribution is a stack attention layer with a focus distance embedding, allowing efficient information exchange across the focus stack. Finally, we develop a new training data pipeline allowing us to utilize existing large-scale RGBD datasets to generate synthetic focus stacks. Experiment results on \benchmarkname\ and other benchmarks show a significant improvement over the baselines, reducing errors by up to 55.7\%. The \benchmarkname\ benchmark is released at \url{https://zedd.cs.princeton.edu}. The code and checkpoints are released at \url{https://github.com/princeton-vl/FOSSA}.
\end{abstract}

\begin{figure}[htbp]
\vspace{-0.5cm}
    \centering
    \includegraphics[width=\textwidth]{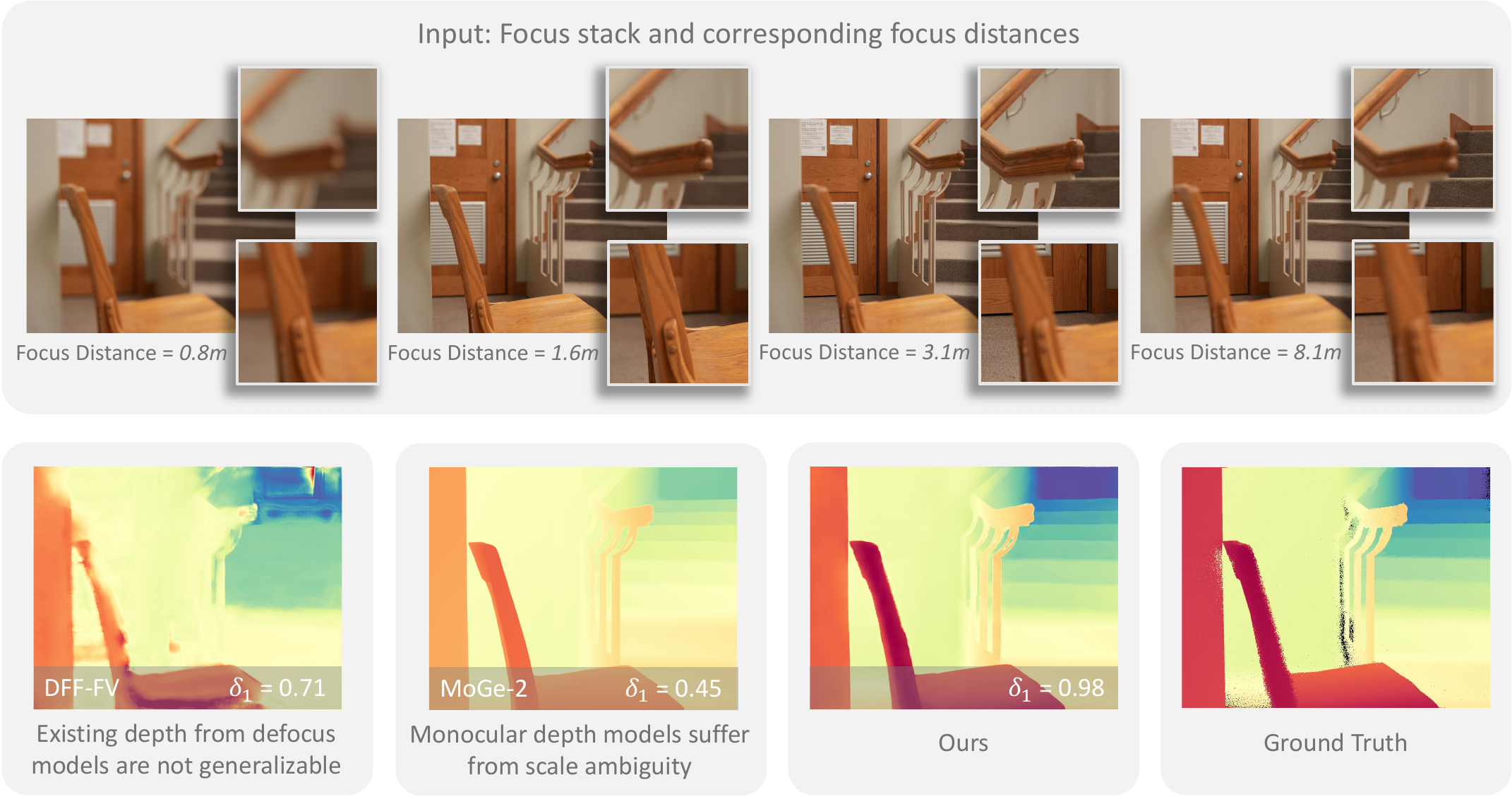}
    \label{fig:teaser}
\vspace{-0.5cm}
\end{figure}

\section{Introduction}
\label{sec:intro}

In modern photography, a technique called focus bracketing is widely adopted. By sweeping the focal plane from near to far, the camera captures a set of images, also known as a \textit{focus stack}. While focus bracketing is traditionally used to create an all-in-focus image, it contains rich information about the scene geometry and depth relationships: as the focal plane sweeps through the scene, objects at different depths sequentially come into focus while others become blurred.

In this paper, we address the task of Depth from Defocus (DfD), which aims at estimating a dense metric depth map from a focus stack. The depth map is useful for various downstream applications, such as post-capture defocus control~\cite{bokehme,mu2025generative}, novel view synthesis~\cite{SHARP,khan2023tiled}, relighting~\cite{yang2021multi}, etc.

While various previous works have tackled the DfD task over the years~\cite{wilddfd,gur2019single,ikoma2021depth,defocusnet,dfv,yang2023aberration,hybriddepth,dualfocus,aifnet,ddff}, they primarily focus on the in-domain setting. As a result, these models do not generalize well to unseen domains, which hinders their application in real-world scenarios. In contrast, we tackle the challenging problem of \textit{zero-shot DfD}, where the models are evaluated on datasets and domains unseen during training. We address two main problems: \textit{1. Lack of high-quality DfD benchmarks} and \textit{2. DfD models not generalizing well}.

\begin{figure}[t]
    \centering
    \vspace{-4mm}
    \includegraphics[width=\textwidth]{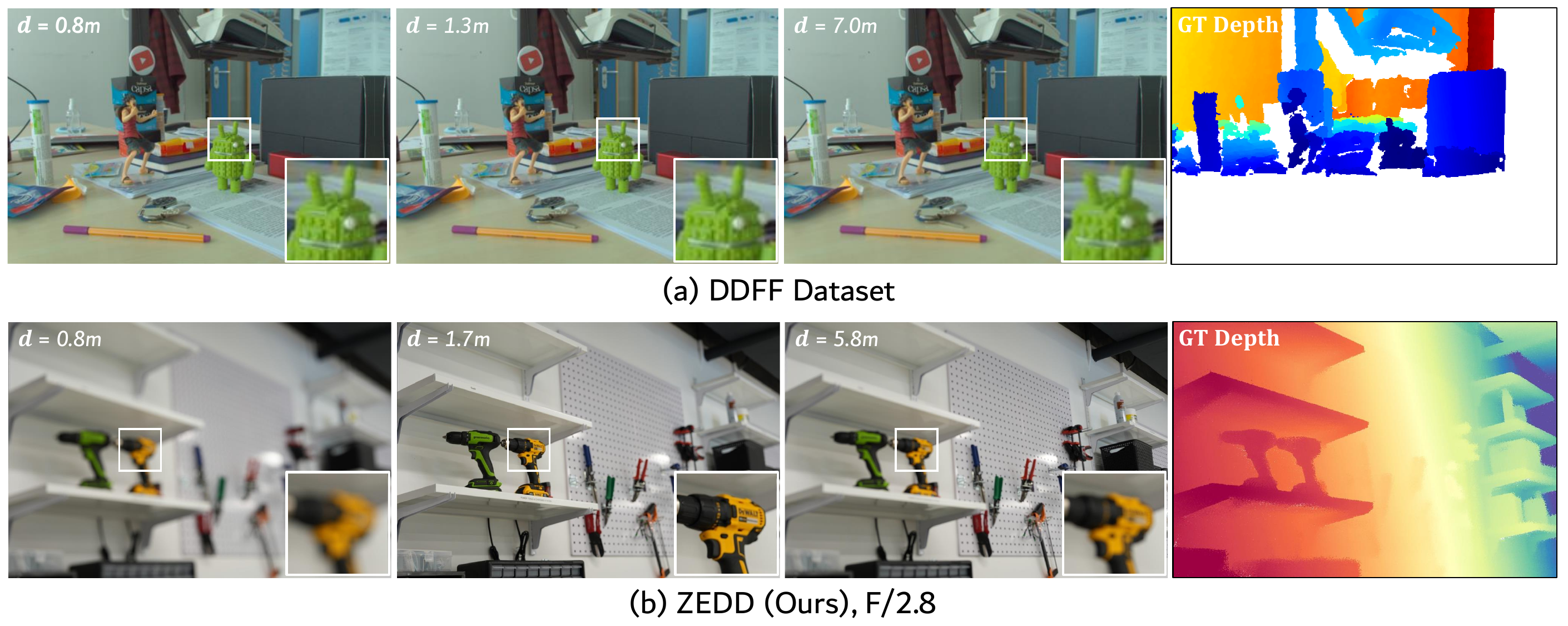}
        \vspace{-7mm}
    \caption{Qualitative comparison between  \benchmarkname\ and DDFF~\cite{ddff}. DDFF uses a very small aperture, so the defocus effect is barely noticeable even when zoomed in. In contrast, our focus stacks exhibit clear, smooth defocus effects. Our depth ground-truth is of higher resolution and denser, and it contains no missing regions due to occlusions.}
    \vspace{-1mm}
    \label{fig:compare_ddff}

\end{figure}

\noindent\textbf{Lack of High Quality Benchmarks.} Previous works~\cite{dualfocus,dered,hybriddepth,dfv} evaluate their models with synthetically generated focus stacks using a 2D point spread functions (PSF) on RGBD datasets such as NYU~\cite{nyuv2}.
However, this 2D processing method only approximates defocus and cannot fully reproduce real optical blur, which depends on 3D scene geometry.
DDFF~\cite{ddff} is the only dataset with real focus stacks and dense depth ground-truth, but has several limitations. Firstly, the focus stack is composed from a light-field image, with the aperture being so small that the defocus is barely noticeable. Secondly, the depth ground-truth is collected from a structured light sensor, which is low-resolution, noisy, and measures only up to 3.5m (\cref{fig:compare_ddff}).

In this paper, we propose a new benchmark \textbf{\benchmarkname}\ (\textbf{ZE}ro-shot \textbf{D}epth from \textbf{D}efocus), a high-quality and diverse real-world benchmark for DfD. We collect images with a 4K resolution DSLR camera under multiple apertures, up to F/1.4. We design a pipeline for accurate calibration and control of the focus distances through remote control of the lens motor position. As for the depth ground-truth, we use a high-end Ouster Lidar and accumulate points through time, achieving high accuracy and dense spatial coverage. Finally, our dataset has high diversity: we collect 100 unique scenes, an order of magnitude greater than previous datasets. Detailed comparisons can be found in \cref{tab:dataset_comparison}.

\noindent\textbf{DfD Models Not Generalizing Well.} Previous works~\cite{dualfocus,hybriddepth,dfv} follow the paradigm of \textit{in-domain} evaluation, \ie, the models are trained and tested on the same datasets, often with limited scales (\eg, both DDFF~\cite{ddff} and NYU~\cite{nyuv2} have < 1000 samples). Moreover, their network designs lack sufficient capacity and scalability, preventing them from fully leveraging large-scale training.

We propose a modern network architecture \textbf{FOSSA} (\textbf{FO}cu\textbf{S} \textbf{S}tack \textbf{A}ttention Transformer) for zero-shot DfD. It is based on the powerful Vision Transformer (ViT)~\cite{ViT} but with novel designs tailored to the DfD task. We propose a \textit{stack attention layer} with a special focus distance embedding, which enables efficient extraction of the defocus cues across the focus stack. We then fuse the individual focus stack features into a global feature map at an early stage, effectively reducing the computation cost of the subsequent layers.

Due to the lack of large-scale focus stack datasets, we train fully on synthetically generated focus stacks, allowing us to utilize any existing RGBD datasets. To achieve better generalization ability, we randomize parameters such as focus distance, aperture size, and the shape of the point spread function (PSF).

We evaluate \methodname\ extensively on diverse benchmarks, including \benchmarkname, DDFF~\cite{ddff}, and multiple real-world RGBD datasets. \methodname\ significantly outperforms the baselines. It reduces AbsRel by 55.7\% compared to the second-best method DepthPro~\cite{depthpro} on \benchmarkname, and MSE by 40.4\% on DDFF compared to previous state-of-the-art DualFocus~\cite {dualfocus}. We also demonstrate that \methodname\ is robust to various factors, including the aperture size and the focus stack size.

\begin{table*}[t]
\centering
  \resizebox{\linewidth}{!}{
\begin{tabularx}{1.4\linewidth}{lCCCCCCCCC}
\toprule
\makecell{Datasets}
& \makecell{\# Scenes}
& \makecell{Ground\\Truth}
& \makecell{Image\\Resolution}
& \makecell{Defocus\\Effect}
& \makecell{Aperture}
& \makecell{Focus\\Distances}
& \makecell{\# Images in \\ Focus Stack} \\
\midrule
\zerotoprule
\makecell{MobileDepth~\cite{mobiledepth}} & \makecell{11} & \makecell{\textcolor{gray}{No GT}} & \makecell{$640\times360$} & \makecell{Real} & \makecell{--} & \makecell{--} & \makecell{10} \\
\zerotoprule
\makecell{DDFF~\cite{ddff}} & \makecell{12} & \makecell{Structured \\ Light} & \makecell{$552 \times 383$} & \makecell{Synthetic \\(Light Field)} & \makecell{F15.0*} & \makecell{0.5~-~7.0m} & \makecell{10} \\ 
\zerotoprule
\makecell{\textbf{\benchmarkname\ (Ours)}} & \makecell{100} & \makecell{Lidar} & \makecell{$4104\times2736$} & \makecell{Real} & \makecell{F1.4/2.0/2.8/ \\ 4.0/5.6/16.0} & \makecell{0.8~-~8.1m} & \makecell{9} \\
\bottomrule
\end{tabularx}
}
\vspace{0.05em}
\caption{Compared to existing DfD benchmarks MobileDepth~\cite{mobiledepth} and DDFF~\cite{ddff}, our \benchmarkname\ has $8.3\times$ more scenes, higher quality ground-truth, higher resolution images, and realistic defocus effect under multiple f-numbers. *: full-frame equivalent aperture computed from specifications of the Lytro camera.}
\label{tab:dataset_comparison}
\vspace{-9mm}
\end{table*}

\section{Related Works}
\subsection{Depth from Defocus Methods}

Recent deep-learning-based methods focus on advancing network designs. DefocusNet~\cite{defocusnet} processes each image with individual branches and fuses information through repetitive pooling. DFV~\cite{dfv} proposes a network based on differential focus volume, which explicitly compares the features of neighboring frames in a focus stack. HybridDepth~\cite{hybriddepth} proposes a multi-stage framework that first aligns relative monocular depth to metric DfD depth, and then performs refinements. DualFocus~\cite{dualfocus} introduces a variational approach that predicts gradient constraints and integrates them into depth maps with a least-squares solver. DEReD~\cite{dered} explores the fully self-supervised setting, trained through a differentiable layer for synthesized defocus effect and photometric reconstruction. Unlike DDFS~\cite{ddfs}, which carefully models and embeds full camera settings, we only embed focus distances as metric anchors, but maintain robustness to focal length and aperture by training the model on diverse combinations.

Compared to previous works, our method has two unique benefits. \textit{1. Simple and Scalable:} previous works are built heavily on custom network layers and designs, making it hard for them to scale up. Comparatively, \methodname\ is simple yet effective: it is based on Vision Transformer (ViT) and is built with standard transformer blocks, which is easy to scale up and benefit from the powerful pretrained models such as DepthAnything~\cite{DAv2}. \textit{2. Generalizable:} While previous works achieve good accuracy on dedicated benchmarks, they do not generalize well beyond certain datasets, and require separate data and training for each benchmark. In comparison, \methodname\ has a single set of weights that works well across all datasets in a zero-shot manner, thanks to our scalable model design (\cref{sec:network_architecture}) and data pipeline (\cref{sec:methods_training}) aiming for maximum generalization.

\subsection{Depth from Defocus Datasets}

Defocus effects are common in real photos. Multiple datasets have been collected for depth estimation on those images. iDFD~\cite{nazir2023idfd}, D3Net~\cite{d3net}, and Talegaonkar \etal~\cite{talegaonkar2025repurposing} collect scenes with paired wide and narrow aperture images, but at the same focus distance, therefore not suitable for the DfD task. 

Several real-world datasets have been collected specifically for DfD, but they are all limited in their scale and quality. The MobileDepth dataset~\cite{mobiledepth} has only 11 scenes captured with a mobile phone, with no depth ground-truth available for quantitative evaluation. DDFF~\cite{ddff} is composed of 720 focus stacks from only 12 scenes, therefore has high content overlap among focus stacks. Moreover, the focus stacks are not realistic: the images are not from a regular lens but synthesized from a light-field camera. Finally, the ground-truth is collected from a structured light sensor, which has limited accuracy, FoV, and range coverage. Compared to DDFF~\cite{ddff}, our dataset has higher diversity (100 distinct scenes), higher quality defocus images (4K resolution with multiple apertures up to F/1.4), and more accurate ground-truth (from a high-end Ouster Lidar).

Others use graphics engines to render synthetic focus stack images. FoD500~\cite{defocusnet} creates toy-like scenes in Blender with randomly scattered objects, and renders defocused images at $256 \times 256$ resolution. We propose a new synthetic benchmark (\cref{sec:exp_datasets}) based on Infinigen~\cite{infinigen-indoors,infinigen-nature} adopting a similar rendering strategy, but adopt procedural scene generation for systematic layout control and achieve more realistic scene structures, higher photorealism, and higher resolution.

\section{Methods}
\subsection{Problem Definition}

The input to the task of Depth from Defocus (DfD) is a focus stack and the corresponding focus metadata. Specifically, the focus stack is a set of $M$ pixel-aligned RGB images $\mathbf{I} = (\mathbf{I}_1,\dots,\mathbf{I}_M) \in \mathbb{R}^{M \times 3 \times H \times W}$, captured from a static scene using a fixed viewpoint and the same aperture, but with different focus distances. The focus metadata is the focus distances $\mathbf{d} = (d_1,\dots,d_M) \in \mathbb{R}^{M}$ in meters. The goal is to predict a dense metric depth map
$\mathbf{\hat{D}} \in \mathbb{R}^{H \times W}$.

\begin{figure}[t]
    \centering
    \includegraphics[width=\textwidth]{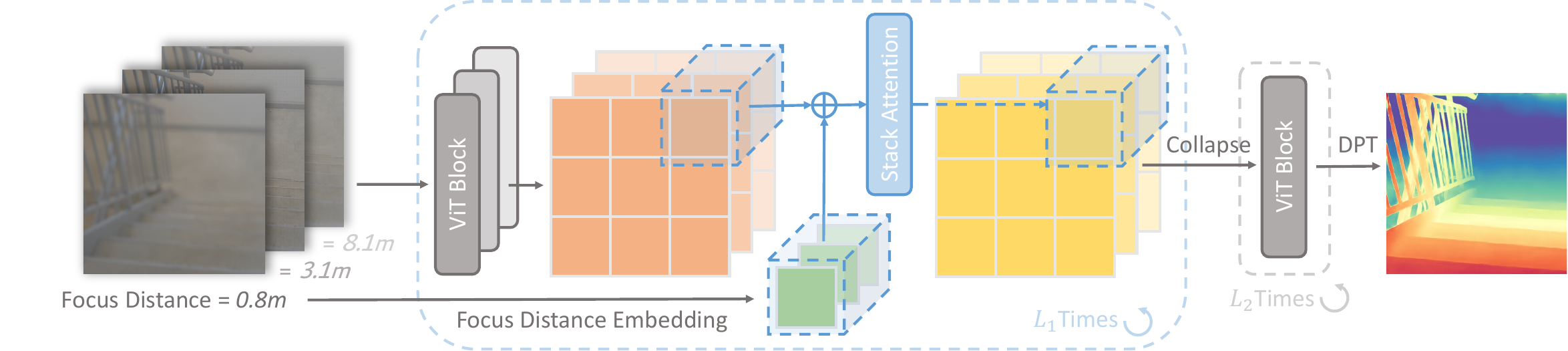}
    \vspace{-3mm}
    \caption{Overview of the \methodname\ pipeline. \methodname\ consists of two main stages: focus stack feature extraction (\textcolor{cyan}{blue} box), and feature refinement (\textcolor{gray}{gray} box). For each focus stack extraction layer, the image features are first processed individually and then pass through a stack attention layer for efficient information exchange across the stack. The features are then collapsed along the stack dimension, and pass through a sequence of refinement ViT blocks and finally a DPT head for dense depth output.}
    \label{fig:pipeline}
    \vspace{-3mm}
\end{figure}

\subsection{\methodname\ Network Architecture}
\label{sec:network_architecture}
As shown in \cref{fig:pipeline}, \methodname\ consists of two stages. It first extracts a per-image feature map for all images in the focus stack, then collapses them into a single global representation for depth prediction. In the first stage, the network produces a feature map for each image in the focus stack using a feature extractor containing $L_1$ number of layers. Each layer contains (i) a weight-shared Vision Transformer (ViT)~\cite{ViT} block applied to each focus image individually and (ii) a novel stack attention layer that enables information exchange across the focus stack dimension.
In the second stage, we collapse the per-focus feature maps into one global feature map, which is further refined by additional $L_2$ ViT blocks and finally fed to a DPT~\cite{DPT} head to regress a dense metric depth map. 

\noindent\textbf{ViT Blocks.} We adopt standard ViT~\cite{ViT} blocks as the backbone for feature encoding. ViT has become a strong backbone for depth estimation and is widely adopted in prior work~\cite{DAv2, videoDA, vggt}.  A ViT block contains multi-head self-attention followed by a feed-forward MLP, with residual connections and normalization.

\noindent\textbf{Focus Stack Feature Extraction.} For each  image $\mathbf{I}_i \in \mathbb{R}^{3 \times H \times W}$ in the stack, we partition it into $p \times p$ patches and embed each patch as a token. This yields a patch-level feature map $\mathbf{F}_i \in \mathbb{R}^{C\times (\frac{H}{p} \times \frac{W}{p})}$, where $C$ is the feature dimension. 

A na\"ive way is to process each $\mathbf{F}_i$ individually using a sequence of ViT blocks to obtain an image feature map, as in monocular depth estimation. In the DfD task setting, however, the key geometric signal is not contained in any single image, but across the focus stack.
Consider a certain patch in the focus stack: as the focus distance sweeps from near to far, the patch turns from blurred to sharp when the focus distance approaches its depth, and becomes blurred again as the focus distance further increases. The visual variation, when interpreted together with the focus distances $\mathbf{d}$, provides a strong cue for metric depth. Exploiting this structure requires explicit information exchange across the focus stack.

To this end, we introduce a stack attention layer and insert it between each ViT block. For each image $\mathbf{I}_i$, we embed its focus distance $d_i$ using a two-layer MLP to obtain a $C$-dimensional vector, and add it to the image tokens. We then concatenate the feature maps across the focus stack to form a 4D feature map $\mathbf{F}^{\prime} \in \mathbb{R}^{M \times C\times (\frac{H}{p} \times \frac{W}{p})}$ and apply attention along the first (stack) dimension. For each patch location, the module performs self-attention across the $M$ focus settings, enabling the network to aggregate defocus cues across the stack.

This design is computationally efficient: the quadratic attention cost is incurred only over the stack size $M$, which is much smaller than the number of spatial tokens, providing effective and efficient cross-image interaction.

\noindent\textbf{Feature Collapse and Refinement.}
After extracting $M$ per-focus feature maps, we collapse them into one global feature map by averaging the features over the focus stack dimension. The global feature has the same shape as a regular image feature ($\mathbb{R}^{C\times (\frac{H}{p} \times \frac{W}{p})}$), and is then refined by subsequent $L_2$ ViT blocks and a standard DPT decoder to produce final depth prediction $\hat{\mathbf{D}}$.

\subsection{Training}
\label{sec:methods_training}

The main challenge for training a zero-shot DfD model is that there are no large-scale DfD datasets that can support training. Therefore, we train our model on existing RGBD datasets using focus stacks generated with synthetic blur. 

The strength of blur depends on multiple factors, including the focal length, aperture, focus distance, and object distance. This strength is often described using the circle of confusion (CoC). Given an image $\mathbf{I}$ with ground truth depth $\mathbf{D}$, the CoC is calculated as:
\begin{equation}
    \text{CoC} = \frac{|\mathbf{D} - d|}{\mathbf{D}} \frac{f^2}{N(d-f)},
\end{equation}
where $d$ is the focus distance, $f$ is the focal length, and $N$ is the f-number. Based on CoC, we compute a point spread function (PSF) as a pixel-wise blur kernel, which is applied to $\mathbf{I}$ through convolution. Unlike prior work that assumes a Gaussian PSF~\cite{dered,gur2019single}, we randomize the PSF shape to better reflect real imaging physics, which varies between diffraction-limited and defocus-dominated regimes. We also randomize the f-number and the focus distance distribution for better generalization. More details can be found in Appendix \cref{sec:appendix_training_details}.

The loss is a combination of the SiLog loss~\cite{silog} with $\lambda=0.5$ to balance metric and relative supervisions, and the Gradient Matching loss~\cite{midas}: 

\begin{equation}
\label{eqn:loss}
    L(\mathbf{\hat{D}, \mathbf{D}}) = \mathrm{SiLog}(\mathbf{\hat{D}, \mathbf{D}}) + 0.1 \cdot\mathrm{GradMatching(\mathbf{\hat{D}, \mathbf{D}})}. 
\end{equation}

\section{\benchmarkname: Real-World DfD Benchmark}

\begin{figure}[t]
    \centering
    \includegraphics[width=\textwidth]{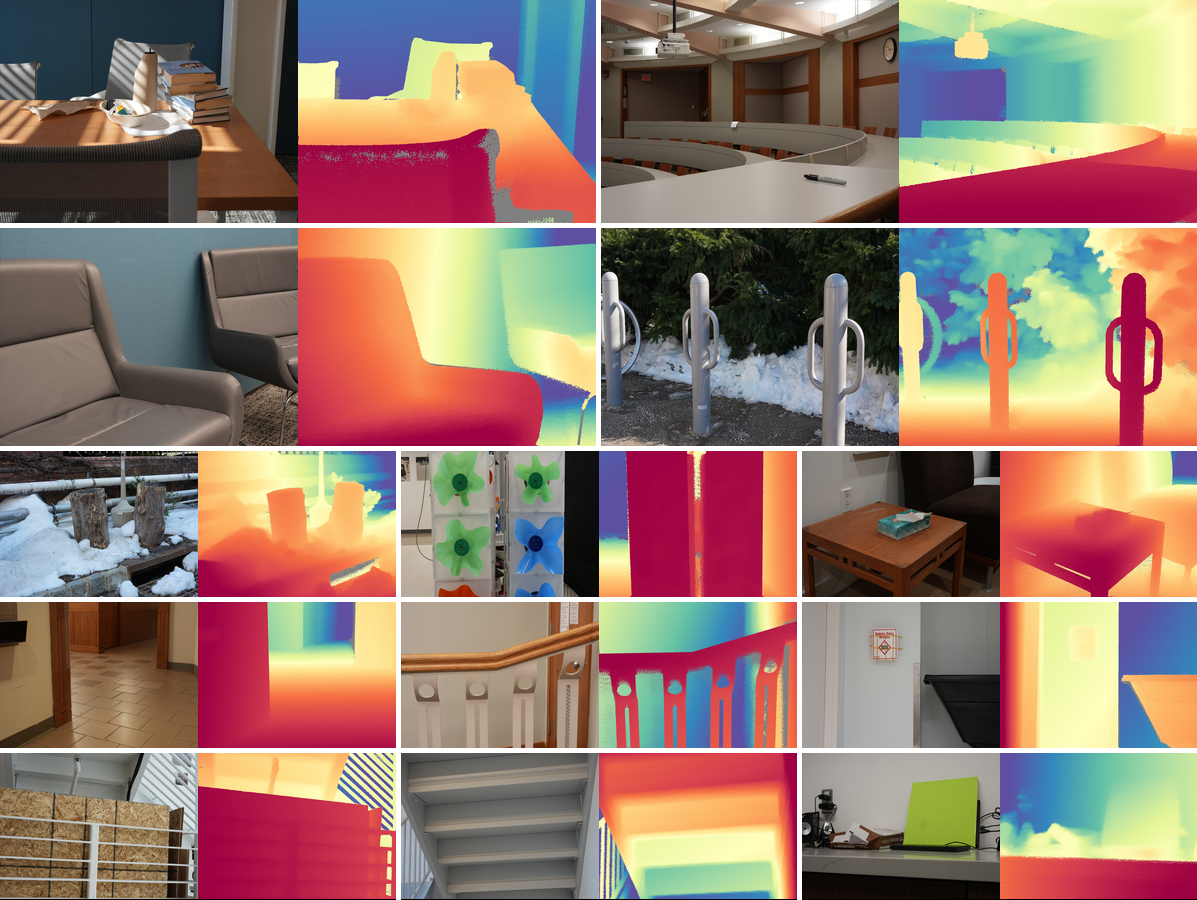}
    \vspace{-3mm}
    \caption{A gallery of our \benchmarkname\ benchmark. \benchmarkname\ includes 100 diverse scenes spanning a wide range of indoor and outdoor environments. The dataset features rich geometric structure and provides high-quality ground-truth.}
    \label{fig:gallery}
\end{figure}

\subsection{Scene Coverage and Statistics}

We captured 100 focus stacks in 100 unique scenes, covering various indoor and outdoor locations, such as classrooms, hallways, robotics labs, offices, kitchens, and gardens, providing a diverse scene coverage.

For each focus stack, we capture images at 9 focus distances, ranging from 0.82 to 8.10m. We capture at 5 larger apertures (F1.4/2.0/2.8/4.0/5.6), and a small aperture (F16) for all-in-focus images, resulting in $6\times9=54$ images in total for each scene. This rich combination of focus distances and apertures allows us to study the sensitivity of the models' performance to each factor (\cref{sec:exp_robustness}).

We provide a dense ground-truth depth map for each scene under the resolution of $1824\times1216$, captured with a high-accuracy Lidar.

\subsection{Focus Stack Capture Procedure}

\textbf{Hardware.} We capture images using a DSLR camera (Sony $\alpha1$-II) together with a prime lens with a large and variable aperture (Sony FE 50mm F1.4 GM). This combination allows us to capture high-resolution 4K images with multiple defocus strengths.

\noindent\textbf{Intrinsics Calibration.} Our goal is to get accurate camera intrinsics. However, the FoV changes as we adjust the focus distance, a phenomenon known as lens breathing. To resolve this, we define a canonical image space, where we set the lens focus distance to be 3.08m. The intrinsics for the canonical space is calibrated using Kalibr~\cite{kalibr}, and images captured under all other focus distances are warped into the canonical space using an open-source focus stacking toolbox~\cite{focusstacktoolbox}.

\noindent\textbf{Focus Distance Calibration.} The autofocus lens we use allows us to read out the focus motor positions as a 4-digit hexcode. The hexcode is a 1-to-1 mapping to the actual focus distance, but this mapping is lens-specific and unknown. Therefore, we recover this mapping as a sparse but accurate lookup table through calibration. Specifically, we place a calibration board parallel to the camera, manually adjust the motor until the board is sharply in focus, and capture an image ($I_0$). We then fix the motor and record the hexcode, and move the board around to capture multiple images ($I_{1 \dots N}$). We finally co-calibrate the intrinsics and extrinsics for all images ($I_{0 \dots N}$) using OpenCV \cite{opencv_library}. The $z$ component of the extrinsics of $I_0$ is used as the ground-truth focus distance.

\noindent\textbf{Software Control.} We capture the focus stack through software control of the camera, providing two main benefits: 1. The software allows us to set the focus motor precisely at the hexcode we have calibrated, making the image-to-focus-distance mapping accurately reproducible. 2. This eliminates possible mechanical vibrations during the capture, so the images in the stack are perfectly aligned.

\subsection{Depth Ground-Truth Acquisition}
\noindent\textbf{Hardware.} We use the Ouster OS0-128 Lidar, which provides a range of up to 100m with a sub-centimeter range accuracy. We select this Lidar for its longer ranges and higher accuracy compared to other commonly used depth sensors such as Intel RealSense.

\noindent\textbf{Point Accumulation.} OS0-128 has a fixed sparse scanning pattern with 128 distributed layers. To achieve a dense depth coverage, we handhold the lidar to move it and accumulate the points for about 600 frames. The frames are registered using an ICP algorithm~\cite{icp} with Open3D~\cite{open3d} with noise reduction. The accumulated point cloud is denoted as $\mathbf{P}_{world}$. 

\noindent\textbf{Registration.} Once a dense ground truth per scene $\mathbf{P}_{world}$ was captured, we aim to transfer it to the camera frame. To achieve this, we need to estimate the projection matrix $^{cam}\mathbf{T}_{world}$ from the world coordinate to the camera coordinate. The camera and the Lidar are rigidly mounted on a tripod, related by a transform $^{cam}\mathbf{T}_{lidar} \in \mathrm{SE}(3)$, which is calibrated separately offline. Therefore, it suffices to localize the Lidar on the tripod in the world coordinate frame. To this end, we capture at every camera location a single Lidar point cloud $\mathbf{P}_{lidar}$ and solve for $^{lidar}\mathbf{T}_{world}$ using ICP between $\mathbf{P}_{lidar}$ and $\mathbf{P}_{world}$. Finally, the camera-space point cloud is computed as $\mathbf{P}_{cam} = {}^{cam}\mathbf{T}_{lidar} {}^{lidar}\mathbf{T}_{world} \mathbf{P}_{world}$.

\noindent\textbf{Projection.} We project $\mathbf{P}_{cam}$ into a depth map with z-buffering~\cite{zbuffer} using the calibrated intrinsics. Note, the projected pointcloud $\mathbf{P}_{cam}$ is dense (up to 1080p resolution) and does not have missing regions due to the parallax effect problem as in many other RGBD datasets~\cite{ddff,qiu2019deeplidar}. We finally do manual quality checks and depth map clean-ups, as detailed in the Appendix \cref{sec:Aggregation}. 

\section{Experiments}

\subsection{Implementation Details}
\label{sec:implementation_details}

We implement \methodname\ with 2 different sizes (ViT-S, ViT-B). We use $L_1=4$ focus stack feature extraction layers and $L_2=8$ refinement ViT layers. We initialize the ViT blocks from the pretrained DepthAnythingv2~\cite{DAv2} indoor metric checkpoint, and zero-initialize the MLP in the stack attention layers. We use the CSTM-label~\cite{metric3dv1,metric3dv2,depthpro} normalization using ground-truth focal length and predict in the normalized depth space. We provide the ground-truth focal length to both our method and baselines~\cite{depthpro,moge2,unidepthv2} for fair comparison.

We train \methodname\ on a combined dataset of Hypersim~\cite{hypersim} (66k samples) for indoor coverage and TartanAir~\cite{tartanair} (307k samples) for outdoor coverage, using synthetically generated focus stacks. We train the model for 40 epochs using an exponential learning rate scheduler, with a batch size of 8. Training takes 2 days on 4 $\times$ L40 GPUs (48GB memory).

We train with a stack size of 5 images. We randomize the near and far bounds of the focus distances, as well as how the other focus distances are interpolated in between. We also randomize the blur kernel by altering the PSF shape and the f-number. See more details in Appendix \cref{sec:appendix_training_details}.

\subsection{Datasets and Evaluation Metrics}
\label{sec:exp_datasets}

\noindent\textbf{\benchmarkname.} Our real-world dataset is the main evaluation benchmark. We randomly divide it into a validation and a test split, with 50 samples in each split. Out of the 9 focus distances (FD), we subsample with stride 2 and create a focus stack of 5 images, with FD=$\{0.8,1.7,3.1,4.9,8.1\}m$, and aperture=F/2.8. 

\noindent\textbf{Infinigen Defocus.} We create a synthetic benchmark based on Infinigen Indoors~\cite{infinigen-indoors}, and we render the focus stacks with defocus effect created using the ray tracer (Cycles) in Blender \cite{blender}. This benchmark is a digital counterpart of the real dataset, which is less photorealistic but has perfect ground-truth. It contains 200 scenes in total, with FD=$\{0.8,1.7,3.0,4.7,8.0\}m$, and aperture=F/1.4.

\noindent\textbf{Real-World RGBD Datasets.} We further evaluate on 3 real-world RGBD datasets for broader scene coverage of indoors (iBims~\cite{ibims} and DIODE~\cite{diode}), outdoors (DIODE~\cite{diode}), and transparent objects (HAMMER~\cite{hammer}). We simulate the focus stack from the ground-truth depth map as described in~\cref{sec:methods_training}. We sample the focus distances depending on the scene-specific depth statistics, to mimic the real-world scenario where the photographer chooses areas to focus on. We set the aperture to F/1.4. See more details in Appendix \cref{sec:appendix_datasets_and_baselines}.

\noindent\textbf{DDFF~\cite{ddff}.} Although the DDFF~\cite{ddff} dataset has several limitations, we still report the numbers for an easy and thorough comparison with DfD baselines. The data split and evaluation metrics follow previous works~\cite{dualfocus,hybriddepth,dfv,defocusnet}.

\begin{table*}[t]
  \centering
  \resizebox{\linewidth}{!}{
    \begin{tabularx}{1.2\textwidth}{lCCCCCC|CCC}
    \toprule
    \multirow{2}{*}{Methods}
    & \multicolumn{6}{c|}{\benchmarkname\ (Test Split)}
    & \multicolumn{3}{c}{Infinigen Defocus} \\
    
    & $\delta_{1.05}\uparrow$ & $\delta_{1.25}\uparrow$ & $\mathrm{AbsRel}\downarrow$  & MAE $\downarrow$ & MSE $\downarrow$ & RMSE $\downarrow$ 
    & $\delta_{1.05} \uparrow$ & $\delta_{1.25} \uparrow$ & $\mathrm{AbsRel} \downarrow$ \\

    \lightbottomrule
    \rowcolor{gray!20}
    \multicolumn{10}{c}{\rule{0pt}{2.4ex}\textit{Monocular Depth}\rule[-1.3ex]{0pt}{0pt}} \\
    \zerotoprule
    
    DAv2~\cite{DAv2}  & 0.037 & 0.261 & 0.337 & 1.211 & 4.295 & 1.353 & 0.026 & 0.133 & 0.429 \\
    DepthPro~\cite{depthpro} & 0.182 & 0.665 & 0.201 & 0.684 & 2.020 & 0.812 & 0.174 & 0.692 & 0.176  \\
    UniDepthv2~\cite{unidepthv2} & 0.106 & 0.605 & 0.253 & 0.866 & 2.439 & 0.976 &  0.103 & 0.464 & 0.370 \\
    MoGe-2~\cite{moge2} & 0.149 & 0.580 & 0.278 & 0.775 & 1.768 & 0.908 & 0.175 & 0.621 & 0.233\\
    
    \lightbottomrule
    \rowcolor{gray!20}
    \multicolumn{10}{c}{\rule{0pt}{2.4ex}\textit{Depth from Defocus}\rule[-1.3ex]{0pt}{0pt}} \\
    \zerotoprule
    
    DFF-FV~\cite{dfv}  & 0.153 & 0.576 & 0.369 & 0.868 & 2.898 & 1.270 & 0.045 & 0.172 & 0.629 \\
    DFF-DFV~\cite{dfv}  & 0.152 & 0.546 & 0.573 & 1.122 & 4.515 & 1.619 & 0.053 & 0.205 & 0.568 \\
    DEReD~\cite{dered} & 0.128 & 0.505 & 0.347 & 1.056 & 4.077 & 1.368 & 0.108 & 0.446 & 0.355 \\
    HybridDepth~\cite{hybriddepth}  & 0.075 & 0.347 & 0.754 & 1.582 & 6.024 & 2.060 & 0.037 & 0.157	& 1.232 \\    
   Ours (ViT-S) & 0.451 & 0.884 & 0.098 & \textBF{0.370} & \textBF{0.661} & \textBF{0.534} & \textBF{0.520} & \textBF{0.940} & \textBF{0.085}  \\
    Ours (ViT-B)  & \textBF{0.505} & \textBF{0.918} & \textBF{0.089} & 0.371 & 0.988 & 0.543 &  0.420 & 0.936 & 0.091 \\
    \bottomrule
    
\end{tabularx}
}
    \vspace{0.2em}
    \caption{Experiment results on \benchmarkname\ and Infinigen Defocus datasets. All methods are tested zero-shot on these benchmarks. Our method \methodname\ outperforms all monocular depth baselines and DfD baselines by a large margin.}
    \vspace{-8mm}
    \label{tab:main_results_real}
\end{table*}

\noindent\textbf{Evaluation Metrics.} We use standard metrics in monocular depth estimation: absolute relative error (AbsRel) and the inlier percentage $\delta_k$, defined as follows:
\vspace{-1em}
\begin{align*}
    \mathrm{AbsRel} :=\frac{1}{N}\sum_{i=1}^{N}\frac{|\hat{d_i} - d_i|}{d_i}, \quad
    \delta_k :=\frac{1}{N}\sum_{i=1}^N\left[\max(\frac{d_i}{\hat{d_i}}, \frac{\hat{d_i}}{d_i}) < k\right] 
\end{align*}
\vspace{-0.5em}

where $N$ is the number of pixels, $d_i$ is the ground-truth, and $\hat{d_i}$ is the prediction. Note $\delta_{1.25}$ is also commonly denoted as $\delta_1$ in the literature. We also report standard depth metrics MAE, MSE, and RMSE on \benchmarkname. See previous literature~\cite{omnidc} for detailed metrics definition.

\subsection{Comparisons with State-of-the-Art Methods}
\label{sec:main_results}

\noindent\textbf{Baselines.} We compare against state-of-the-art monocular depth baselines~\cite{DAv2,depthpro,unidepthv2,moge2} and DfD baselines~\cite{dfv,dered,hybriddepth}. Monocular methods take a single all-in-focus image as input (an F/16 image for \benchmarkname). Note the comparison to monocular methods is not intended as a direct head-to-head comparison, nor do we claim that our method replaces monocular depth. Instead, we aim to position our method relative to widely used depth-estimation baselines. For the DfD baselines providing multiple checkpoints, we report the best-performing one.

\begin{figure}[t!]
    \centering
    \includegraphics[width=\textwidth]{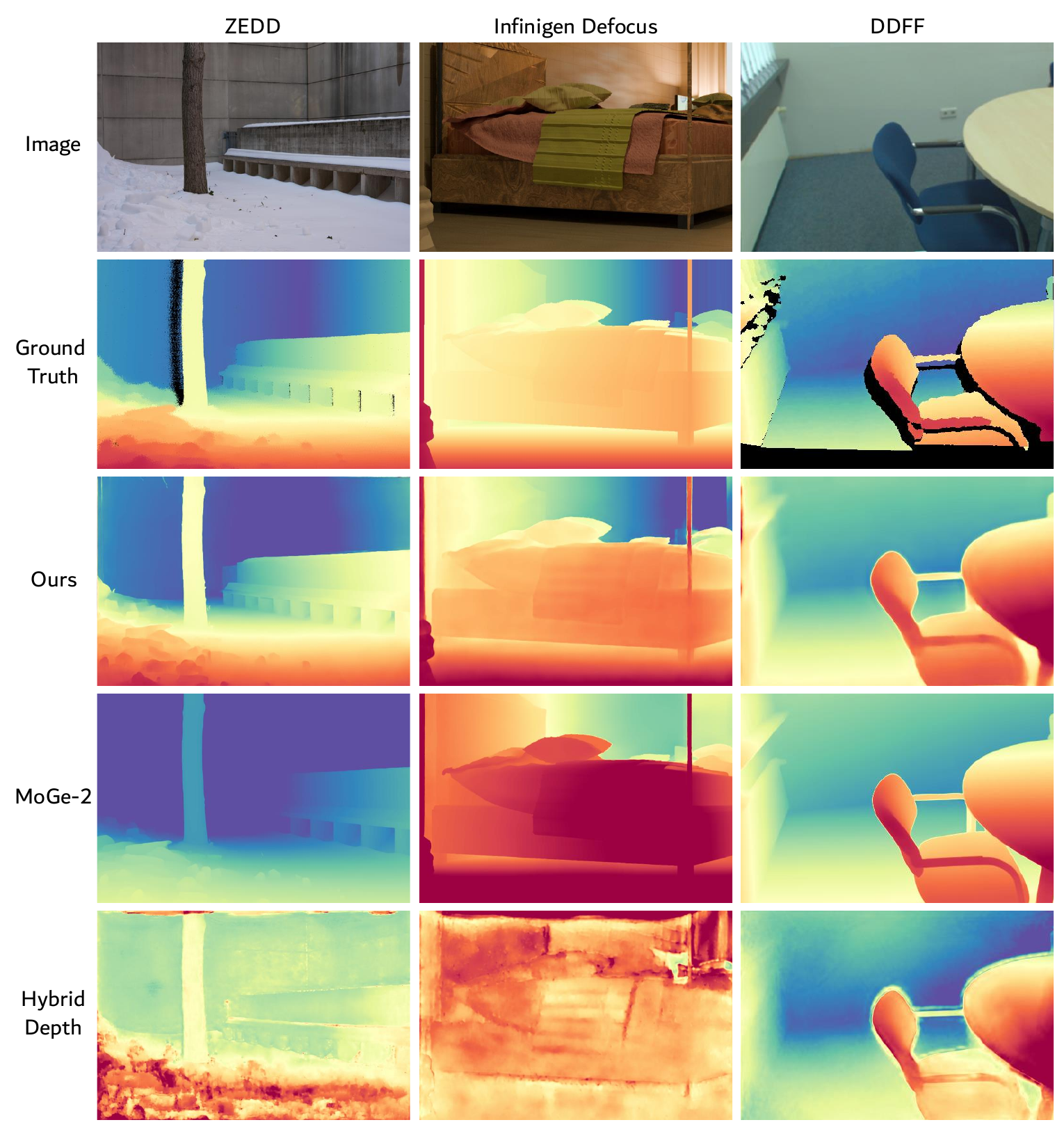}
    \caption{Qualitative comparisons with baselines on 3 benchmarks. Our results are both sharp and metrically accurate. While monocular depth methods such as MoGe-2~\cite{moge2} also produce sharp results, the scales are not metrically correct due to the scale ambiguity. DfD methods such as HybridDepth~\cite{hybriddepth} fail on ZEDD and Infinigen Defocus. Even on DDFF where HybridDepth is trained on, their results are not as sharp as ours.}
    \label{fig:main_results}
    \vspace{-3mm}
\end{figure}

\noindent\textbf{Results on \benchmarkname.} Quantitative results are presented in \cref{tab:main_results_real}. Our model outperforms all monocular and DfD baselines by a large margin, proving the strong effectiveness of our approach. Compared to the best-performing monocular depth baseline DepthPro~\cite{depthpro}, the AbsRel is reduced by 55.7\%, from 0.201 to 0.089. Note that none of the DfD baselines perform well on the \benchmarkname\ benchmark, suggesting their limited generalization ability. The strongest DfD baseline, DFF-FV, is even slightly behind the monocular baselines ($\delta_{1.25}=0.576$ compared to $\delta_{1.25}=0.580$ for MoGe-2~\cite{moge2}). 

Qualitative results are shown in \cref{fig:main_results}. Our results are both sharp and metrically correct. The MoGe-2~\cite{moge2} results are sharp but suffer from scale ambiguity, whereas HybridDepth~\cite{hybriddepth} fails catastrophically due to the large domain gap.

\begin{table*}[t]
  \centering
  \resizebox{\linewidth}{!}{
    \begin{tabularx}{1.1\textwidth}{lCC|CC|CC}

\toprule
\multirow{2}{*}{Methods}
& \multicolumn{2}{c|}{iBims~\cite{ibims}}
& \multicolumn{2}{c|}{DIODE~\cite{diode}} & \multicolumn{2}{c}{HAMMER~\cite{hammer}} \\

& $\delta_{1.25} \uparrow$ & $\mathrm{AbsRel}  \downarrow$
& $\delta_{1.25} \uparrow$ & $\mathrm{AbsRel}  \downarrow$ 
& $\delta_{1.25} \uparrow$ & $\mathrm{AbsRel} \downarrow$ \\

\lightbottomrule
\rowcolor{gray!20}
\multicolumn{7}{c}{\rule{0pt}{2.4ex}\textit{Monocular Depth}\rule[-1.3ex]{0pt}{0pt}} \\
\zerotoprule

DAv2~\cite{DAv2} &  0.886	& 0.127 & 0.440 & 0.312 & 0.320 & 0.541 \\
DepthPro~\cite{depthpro} & 0.878 & 0.120  & 0.462 & 0.289 & 0.719 & 0.272\\
UniDepthV2~\cite{unidepthv2} & 0.941 & 0.078 & 0.680 & 0.368 & 0.482 & 0.371  \\
MoGe-2~\cite{moge2} & 0.886 & 0.107  & 0.757 & 0.341 & 0.755 & 0.266\\

\lightbottomrule
\rowcolor{gray!20}
\multicolumn{7}{c}{\rule{0pt}{2.4ex}\textit{Depth from Defocus}\rule[-1.3ex]{0pt}{0pt}} \\
\zerotoprule

DFF-FV~\cite{dfv} & 0.773 &	0.166 & 0.636 & 0.729 & 0.832 & 0.116\\
DFF-DFV~\cite{dfv} & 0.786 & 0.160 & 0.671 & 0.315 & 0.931 & 0.100\\
DEReD~\cite{dered} & 0.263	& 0.741 & 0.106 & 0.886 & 0.000 & 5.430 \\
HybridDepth~\cite{hybriddepth} & 0.872 & 0.133 & 0.580 & 4.261 & 0.956 & 0.089 \\
DFF-DFV$^\dagger$ &	0.925 & 0.096 & 0.751 & 1.331 & 0.999 & 0.044 \\

Ours (ViT-S) & 0.954 & 0.075 & 0.766 & 0.178 & 0.999 & 0.044 \\
Ours (ViT-B) & \textBF{0.963} & \textBF{0.070} & \textBF{0.779} & \textBF{0.160} & \textBF{0.999}	& \textBF{0.017} \\
\bottomrule
\end{tabularx}
}
\vspace{0.2em}
\caption{Results on 3 real-world RGBD datasets, covering various indoor and outdoor scenes. Focus stacks are synthesized from the ground-truth depth maps using PSF. All methods are zero-shot. $\dagger$ means we re-train the model using the exact same data and loss as ours. Our method \methodname\ outperforms all baselines.}
\label{tab:main_results_ibims}
\vspace{-7mm}
\end{table*}

\noindent\textbf{Infinigen Defocus.} Results are presented in \cref{tab:main_results_real}. Similar to the \benchmarkname\ results, our model outperforms all baselines by a large margin. The AbsRel is reduced by 51.7\% compared to the second-best DepthPro~\cite{depthpro}, from 0.176 to 0.085. Qualitative results are shown in \cref{fig:main_results}. Our model accurately captures thin structures, such as poles, while maintaining the correct global scale.

\noindent\textbf{Real-World RGBD Datasets.} 
We evaluate our model against state-of-the-art methods on iBims~\cite{ibims}, DIODE~\cite{diode}, and HAMMER~\cite{hammer} with synthetic focus stacks. These datasets contain diverse indoor and outdoor scenes, where outdoor scenes are particularly challenging as the defocus effects become weaker as depth values increase. Nevertheless, our method outperforms all baselines across all metrics, as shown in \cref{tab:main_results_ibims}.
Our method reduces the AbsRel by 44.6\% compared to DepthPro~\cite{depthpro} on DIODE~\cite{diode}, from 0.289 to 0.160.

We also fine-tune a DfD baseline DFF-DFV using the same training data as ours. This yields a substantial gain, showing an 11.92\% improvement on $\delta_{1.25}$ on DIODE, highlighting the effectiveness of our synthetic training data generation pipeline. On the other hand, the finetuned DFF-DFV still consistently performs worse than our model. Overall, these results suggest that our improvements arise from both the proposed architecture and the improved data pipeline.

\noindent\textbf{DDFF~\cite{ddff}.} Results are shown in \cref{tab:ddff}. For the finetuned experiments, we finetune using the DDFF training set (400 samples, same as baselines) for another 150 epochs using the same loss as in \cref{eqn:loss}. Thanks to our powerful pretrained weights, the finetuned model significantly outperforms the baselines: the MSE is reduced by 40.4\% compared to the previous state-of-the-art DualFocus~\cite{dualfocus}.

\begin{table*}[t]
  \centering
  \resizebox{\linewidth}{!}{
    \begin{tabularx}{1.2\textwidth}{lcCCCCCC}

\toprule

\multirow{2}{*}{Methods}
& \multicolumn{7}{c}{DDFF Dataset~\cite{ddff}} \\ & MSE $\downarrow(\times 10^{-4})$ & RMSE $\downarrow$ & AbsRel $\downarrow$ & SqRel $\downarrow$ & $\delta_1\uparrow$ & $\delta_2\uparrow$ & $\delta_3\uparrow$ \\

\lightbottomrule
\rowcolor{gray!20}
\multicolumn{8}{c}{\rule{0pt}{2.4ex}\textit{Zero-Shot}\rule[-1.3ex]{0pt}{0pt}} \\
\zerotoprule

Ours (ViT-S, pretrained) & $15.1$  & 0.0352 & 	0.27	& 0.0119	& 0.346	& 0.812	& 0.954 \\
Ours (ViT-B, pretrained) & $12.9$ & 0.0324 & 0.21 & 0.0107 & 0.608 & 0.921 & 0.968 \\

\lightbottomrule
\rowcolor{gray!20}
\multicolumn{8}{c}{\rule{0pt}{2.4ex}\textit{Trained on DDFF}\rule[-1.3ex]{0pt}{0pt}} \\
\zerotoprule

DDFF~\cite{ddff}        & $8.9$  & 0.0276 & 0.24 & 0.0095 & 0.613 & 0.887 & 0.965 \\
DefocusNet~\cite{defocusnet}& $8.6$  & 0.0255 & 0.17 & 0.0060 & 0.726 & 0.942 & 0.979 \\
DFF-DFV~\cite{dfv}        & $5.7$  & 0.0213 & 0.17 & 0.0063 & 0.767 & 0.942 & 0.981 \\
HybridDepth~\cite{hybriddepth} & $5.1$  & 0.0200 & 0.17 & 0.0060 & 0.789 & 0.947 & 0.981 \\
DualFocus~\cite{dualfocus}      & $4.7$ & 0.0194 & 0.16 & 0.0056 & 0.800 & 0.954 & 0.982 \\
Ours (ViT-S, finetuned) & $4.2$ & 0.0183 & \textBF{0.11} & 0.0045 & \textBF{0.936} & 0.983 & 0.991 \\
Ours (ViT-B, finetuned) & $\mathbf{2.8}$ &\textBF{0.0148} & \textBF{0.11} & \textBF{0.0025} & 0.932 & \textBF{0.987} & \textBF{0.994} \\
\bottomrule
\end{tabularx}
}
\vspace{0.2em}
\caption{Results on the DDFF~\cite{ddff} dataset. Numbers are adopted from the DualFocus~\cite{dualfocus} paper. We test under two settings: zero-shot and finetuned on DDFF. Our finetuned checkpoint outperforms all previous DfD baselines by a large margin.}
\label{tab:ddff}
\vspace{-5mm}
\end{table*}

\begin{figure}[!ht]
    \centering
    \includegraphics[width=\textwidth]{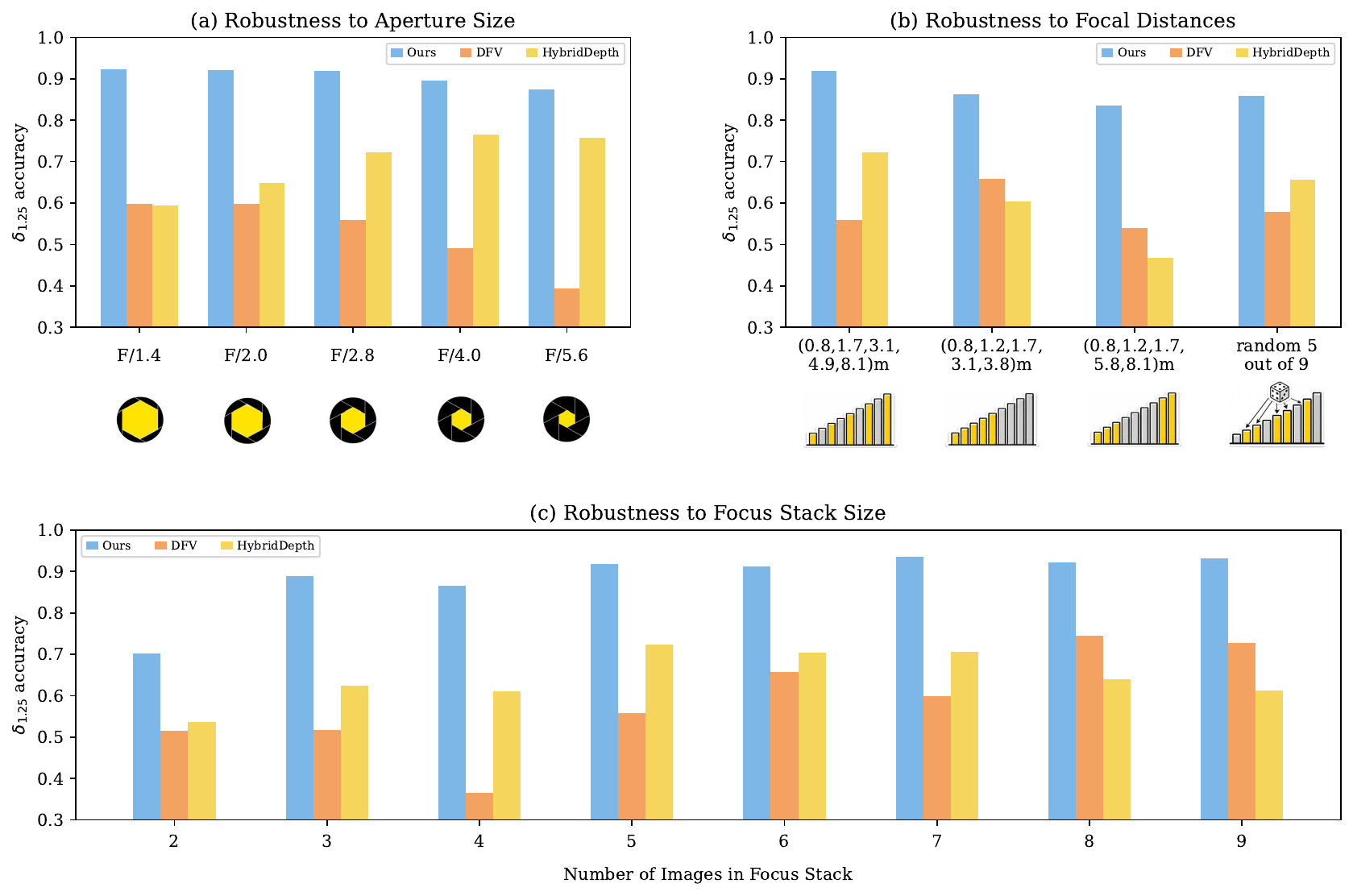}
    \vspace{-4mm}
    \caption{\methodname\ is robust to the aperture sizes, the focus distance distribution, and the focus stack size. The variance is small and we consistently outperform baselines under all configurations. Although always trained with a focus stack size of 5, our method can work with as few as 2 images. All results are on the \benchmarkname\ validation split.}
    \vspace{-5mm}
        \label{fig:robustness}
\end{figure}

\subsection{Robustness}
\label{sec:exp_robustness}

On \benchmarkname\ validation split, we evaluate robustness to three factors: aperture size, the distribution of focus distances, and the number of images in the focus stack. Results are shown in \cref{fig:robustness}.

\noindent\textbf{Aperture size.} Smaller apertures (larger f-numbers) produce weaker defocus blur and are more challenging. As the aperture changes from F/1.4 to F/5.6, the $\delta_{1.25}$ of our method decreases by only 5\%, indicating strong robustness.

\noindent\textbf{Focus distance distribution.} Varying the sampling distribution has a limited impact. Even under an extreme setting diverging largely from the training distribution (the 3rd group, using only the smallest three and largest two focus distances out of nine), $\delta_{1.25}$ drops by only 9\% relative to uniform sampling (the 1st group).

\noindent\textbf{Focus stack size.} Although the model is trained with 5 images per stack, it remains effective with fewer input images and performance degrades only gradually as the stack size decreases. Conversely, providing more than 5 images consistently improves performance.

\subsection{Ablation Studies}

We study the effectiveness of our designs, as presented in \cref{tab:ablations}. Our designs and hyperparameters are general and not tuned to overfit certain datasets, as the same conclusions can be drawn from \benchmarkname\ and Infinigen Defocus.

\noindent\textBF{Stack Attention Layer.} Stack attention and focus distance embedding are crucial to achieving good performance. Removing either component leads to a substantial drop, with $\delta_{1.05}$ decreasing from $0.445$ to around $0.09$.

\noindent\textBF{Where to Collapse.} Collapsing at a later stage (\ie, more focus stack feature extraction layers) brings higher capacity and improves performance, but also higher computational cost. We choose $L_1=4$ to balance accuracy and efficiency. In contrast, fusion within the DPT head such as VideoDA~\cite{videoDA} performs poorly on the DfD task, likely because the DPT head has insufficient capacity to effectively fuse features across the entire focus stack.

\noindent\textBF{Domain Randomization During Training.} Both randomizing the blur kernel and the focus distance distribution are important for robustness, as it reduces overfitting to specific defocus patterns. Removing these two sources of randomization leads to the $\delta_{1.25}$ drops of 21.12\% and 39.78\%, respectively.

\section{Conclusion}

This paper studies zero-shot depth from defocus (DfD), aiming to estimate dense metric depth from a focus stack. We introduce \benchmarkname, a real-world benchmark with broad scene coverage and high-quality ground truth, and propose \methodname, a network built around a stack attention layer. Our method consistently outperforms all DfD and monocular baselines by a substantial margin on various benchmarks. Our paper paves the road for future research on the DfD task.

\begin{table*}[t]
  \centering
  \resizebox{\linewidth}{!}{
    \begin{tabularx}{1.25\textwidth}{lCCCCCCCC}
\toprule
\multirow{2}{*}{Variants}
& \multicolumn{2}{c}{\benchmarkname} & \multicolumn{2}{c}{Infinigen} &  Params & Mem & Speed & \multirow{2}{*}{GFLOPS} \\
& $\delta_{1.05} \uparrow$ & $\mathrm{Rel} \downarrow$ & $\delta_{1.05} \uparrow$ & $\mathrm{Rel} \downarrow$ & (M) 
& (GB) & (ms) &   \\
\lightbottomrule

\rowcolor{gray!20}
\multicolumn{9}{c}{\rule{0pt}{2.4ex}\textit{Stack Attention Layer}\rule[-1.3ex]{0pt}{0pt}} \\
\zerotoprule

1) No stack attention layer & 0.092 & 0.444 & 0.135 & 0.320 & 24.8 & 1.42 & 32.6 & 206.2  \\
2) No FD embedding & 0.090	& 0.399 & 0.123 & 0.365 & 37.8 & 1.52 & 45.7 & 326.7 \\

\lightbottomrule
\rowcolor{gray!20}
\multicolumn{9}{c}{\rule{0pt}{2.4ex}\textit{Where to Collapse}\rule[-1.3ex]{0pt}{0pt}} \\ 
\zerotoprule

3) Collapse at $L_1=2$ & 0.372 & 0.108 & 0.332 & 0.105 & 33.7 & 1.49 & 34.8 & 241.0 \\
4) Collapse at $L_1=6$ & 0.524 & 0.085 & 0.548 & 0.073 & 51.4 & 1.62 & 69.5 & 499.8 \\
5) Fuse in DPT ~\cite{videoDA} & 0.061 & 0.464 & 0.075 & 0.376 & 30.6 & 1.57 & 78.1 & 457.9 \\

\lightbottomrule
\rowcolor{gray!20}
\multicolumn{9}{c}{\rule{0pt}{2.4ex}\textit{Domain Randomization During Training}\rule[-1.3ex]{0pt}{0pt}} \\
\zerotoprule

6) No random blur kernel  & 0.351 & 0.128 & 0.367 & 0.118 & 42.5 & 1.56 & 51.9 & 370.4 \\
7) No random FD distribution  & 0.268 & 0.136 & 0.294 & 0.171 & 42.5 & 1.56 & 51.9 & 370.4 \\

\midrule

\textbf{Ours}  & 0.445 & 0.099 & 0.520 & 0.085 & 42.5 & 1.56 & 51.9 & 370.4 \\
\bottomrule
\end{tabularx}
}
\vspace{0.2em}
\caption{Ablation studies on \benchmarkname\ (Val split) and Infinigen Defocus. All models are ViT-S, under $700\times512$ resolution on a single L40 GPU. Details:  \textbf{2)} Using the regular sinusoidal embedding~\cite{transformer} based on frame index. \textbf{5)} Processing all images in parallel until the end, and fuse the features using stack attention in DPT (design as in VideoDA~\cite{videoDA}). \textbf{6)} Training with F/1.4 and GaussPSF~\cite{dered}. \textbf{7)} Training with scene-dependent focus distances, mimicking a photographer-in-the-loop (Appendix \cref{sec:fd_randomization}).}
\label{tab:ablations}

\vspace{-8mm}

\end{table*}

\section*{Acknowledgments}
This work was partially supported by the National Science Foundation. Felix Heide is a co-founder of Algolux (now Torc Robotics), Head of AI at Torc Robotics, and a co-founder of Cephia AI.

\bibliographystyle{splncs04}
\bibliography{main}

\clearpage

\setcounter{section}{0}
\renewcommand{\thesection}{A\arabic{section}}
\renewcommand{\theHsection}{A\arabic{section}}
\setcounter{table}{0}
\renewcommand{\thetable}{A\arabic{table}}
\renewcommand{\theHtable}{A\arabic{table}}
\setcounter{figure}{0}
\renewcommand{\thefigure}{A\arabic{figure}}
\renewcommand{\theHfigure}{A\arabic{figure}}
\setcounter{algorithm}{0}
\renewcommand{\thealgorithm}{A\arabic{algorithm}}
\renewcommand{\theHalgorithm}{A\arabic{algorithm}}

{\Large\centering\textbf{Appendix}}\\

\section{\benchmarkname\ Data Collection Pipeline}
\subsection{Hardware Setup and Capture Process}

\begin{figure}[!b]
    \centering
\includegraphics[width=\linewidth]{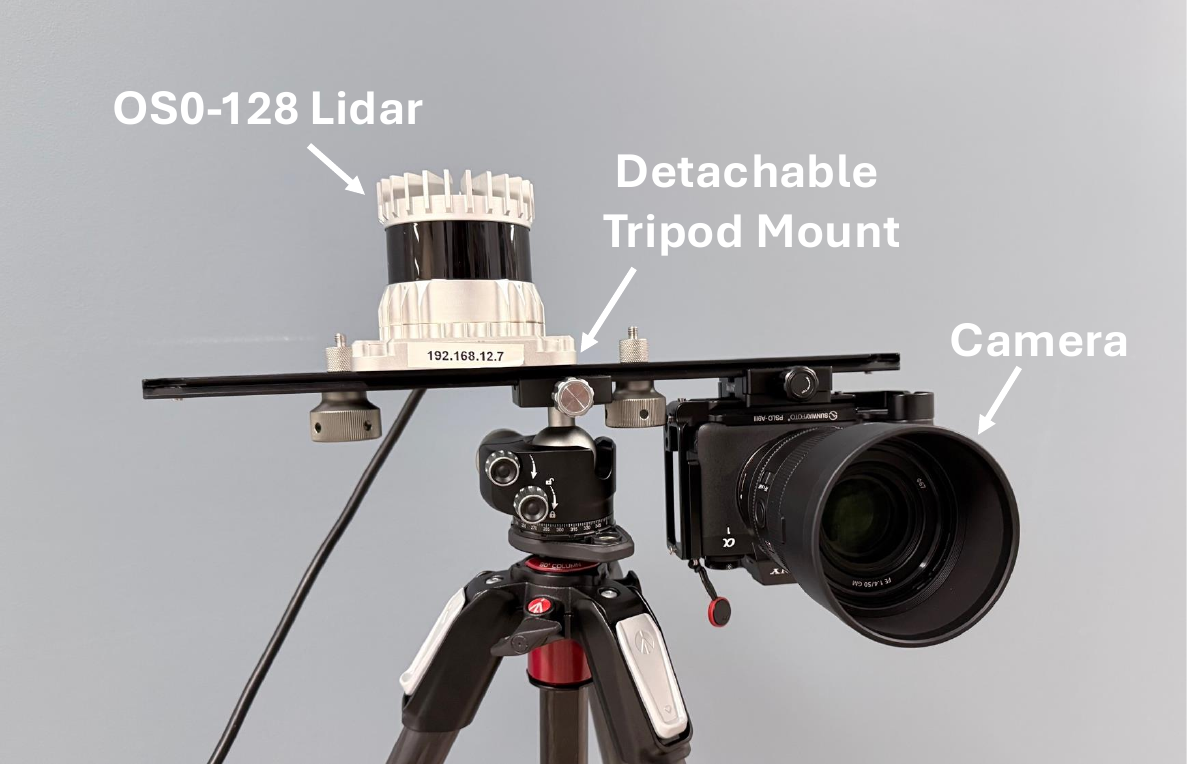}
    \caption{Hardware setup of our data capture pipeline.}
    \label{fig:hardware}
\end{figure}

The hardware setup is shown in \cref{fig:hardware}. The Lidar and the camera are rigidly mounted on a metal bar, so their relative pose does not change during the entire capture process. The mount between the metal bar and the tripod is detachable, so that we can handhold the metal bar to move the Lidar around to get a dense point cloud.

For each scene, the capture process is as follows:
\begin{itemize}
    \item Step 1: We pick a viewpoint and capture a full focus stack through software control of the camera and use the electronic shutter, so no mechanical vibration can cause blur or misalignment. For the best image quality, we adjust the shutter speed together with the aperture, so the ISO is always 100.

    \item Step 2: Without moving anything, we capture a single-frame point cloud with the Lidar, denoted as $\mathbf{P}_{lidar}$. 
    \item Step 3: We remove the metal bar from the tripod, and start moving the Lidar around with our hand. We continuously capture point clouds while moving, at about 10Hz for 60 seconds, resulting in $N \approx 600$ frames $\{\mathbf{P}_i\}_{i=0}^{N-1}$.  
    
\end{itemize}

The capture process takes about 5 minutes for each scene, with about 3.5 minutes for the focus stack capture, and 1.5 minutes for the point cloud capture.

\subsection{Aggregation, Projection, and Post-processing}
\label{sec:Aggregation}

We aggregate $N$ per-frame point maps $\{\mathbf{P}_i\}_{i=0}^{N-1}$ into a single dense global point cloud $\mathbf{P}_{world}$ using the iterative closest point (ICP) \cite{icp} implementation provided by Open3D \cite{open3d}. ICP estimates the rigid transformation between two point clouds by iteratively establishing correspondences and refining the alignment. In our settings, this aggregation procedure is empirically robust enough to handle sensor noise, including scenes containing reflective or transparent materials.

Let $\mathbf{P}_i$ denote the point cloud of the $i$-th frame, and let $\mathbf{G}_i$ denote the global point cloud of the corresponding frame. We initialize the global point cloud as the first frame, i.e., $\mathbf{G}_0 \leftarrow \mathbf{P}_0$. For each subsequent frame $\mathbf{P}_i$, we use ICP to estimate the rigid transformation $^{\mathbf{P}_i}\mathbf{T}_{\mathbf{G}_{i-1}}$, which aligns $\mathbf{G}_{i-1}$ to $\mathbf{P}_i$. We then union the transformed $\mathbf{G}_{i-1}$ together with ${\mathbf{P}_i}$ to get an updated global point cloud, \ie, $\mathbf{G}_{i} \leftarrow ^{\mathbf{P}_i}\mathbf{T}_{\mathbf{G}_{i-1}} (\mathbf{G}_{i-1}) \cup \mathbf{P}_i$. We find that aligning $\mathbf{G}_{i-1}$ to $\mathbf{P}_i$ gives more stable results compared to the other direction, due to the asymmetric nature of the ICP algorithm.

A common failure mode in the aggregated point cloud is the presence of floating points caused by misdetections or imperfect alignment. To suppress such artifacts, we periodically apply a density-based filtering step during aggregation. After a warm-up period of $T$ iterations, we perform the following filtering once every $M$ iterations: for each point $\mathbf{x}$ in the global point cloud, we count the number of neighboring points within a radius $r(\mathbf{x}) = \alpha d(\mathbf{x})$, where $d(\mathbf{x})$ is the distance from $\mathbf{x}$ to the Lidar center, and $\alpha$ is a scale factor. Points with fewer than $k$ neighbors inside this radius are removed. This adaptive radius makes the filtering threshold grow with distance, which is appropriate because the Lidar sampling is uniform in radian space, making the point spacing in Euclidean space proportional to its distance from the sensor. 

The full procedure is summarized in \cref{alg:aggregation}. For all scenes, we set $\alpha = 0.008$, $k = 7$, warm-up steps $T = 50$, and filtering interval $M = 10$.

\begin{algorithm}[t]
\caption{Point-cloud aggregation with outlier filtering}
\label{alg:aggregation}
\begin{algorithmic}[1]
\State \textbf{Input:} Point clouds $\{\mathbf{P}_i\}_{i=0}^{N-1}$; filter warm-up steps $T$; filter interval $M$; neighbor radius threshold $\alpha$; neighbor count threshold $k$.
\State Initialize global point cloud: $\mathbf{G}_0 \gets \mathbf{P}_0$
\For{$i = 1$ to $N-1$}
    \State Estimate $^{\mathbf{P}_i}\mathbf{T}_{\mathbf{G}_{i-1}} \gets \mathrm{ICP}(\mathbf{P}_i, \mathbf{G}_{i-1})$
    \State Transform global point cloud: $\mathbf{G}'_{i-1} \gets ^{\mathbf{P}_i}\mathbf{T}_{\mathbf{G}_{i-1}}(\mathbf{G}_{i-1})$
    \State Merge the aligned frame: $\mathbf{G}_i \gets \mathbf{G}'_{i-1} \cup \mathbf{P}_i$
    \If{$i > T$ and $i \bmod M = 0$}
        \State Remove points in $\mathbf{G}_i$ that have fewer than $k$ neighbors within $r(\mathbf{x}) = \alpha d(\mathbf{x})$
    \EndIf
\EndFor
\State \textbf{Output:} Aggregated global point cloud $\mathbf{G}_{N-1}$
\end{algorithmic}
\end{algorithm}

\begin{figure}[t]
    \centering
\includegraphics[width=\linewidth]{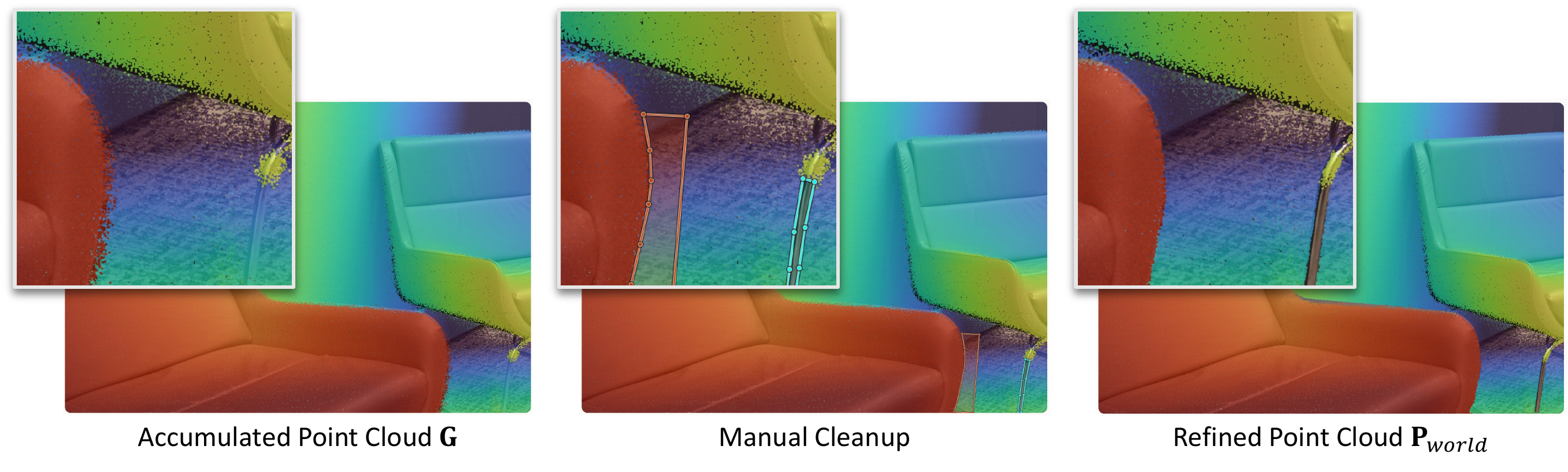}
    \caption{Manual cleanup procedure. In the lower-right corner of the left chair, Lidar noise accumulates near the occlusion boundary, producing an enlarged silhouette (red region). The leg of the right chair is metallic and highly reflective, leading to missing depth measurements (blue region). We manually select and remove these noisy regions.}
    \label{fig:manual_cleanup}
\end{figure}

The automatic filter removes most of the outlier points, but artifacts still remain in the aggregated point cloud. We observe two common remaining failure modes. First, the reconstruction is often noisy near occlusion boundaries. This is because the Lidar measurements typically have centimeter-level depth noise, which can accumulate across frames and produce an artificially enlarged silhouette around depth discontinuities. Second, highly reflective surfaces often lead to unreliable Lidar returns, resulting in severe noise or missing depth. To address these cases, we perform an additional manual cleanup step. We design an interactive UI that allows an annotator to select regions of the aggregated point cloud and remove the points within a certain depth range in that region. We manually inspect and clean all scenes in this manner, and denote the resulting refined global point cloud by $\mathbf{P}_{world}$, as shown in \cref{fig:manual_cleanup}.

After obtaining the refined global point cloud $\mathbf{P}_{world}$, we localize the single-frame Lidar point cloud $\mathbf{P}_{\text{lidar}}$ by registering $\mathbf{P}_{\text{lidar}}$ to $\mathbf{P}_{world}$ using ICP. This yields the relative pose between the current frame and the global reconstruction. We then transform $\mathbf{P}_{world}$ into the coordinate system of $\mathbf{P}_{\text{lidar}}$, and render the transformed point cloud into a dense depth map using z-buffering~\cite{zbuffer}. To account for depth range differences across scenes, we manually select the z-buffer point radius on a per-scene basis, using a value of either $3$ or $4$ pixels.

\subsection{Scene Statistics}
\begin{figure}[!t]
    \centering
\includegraphics[width=0.7\linewidth]{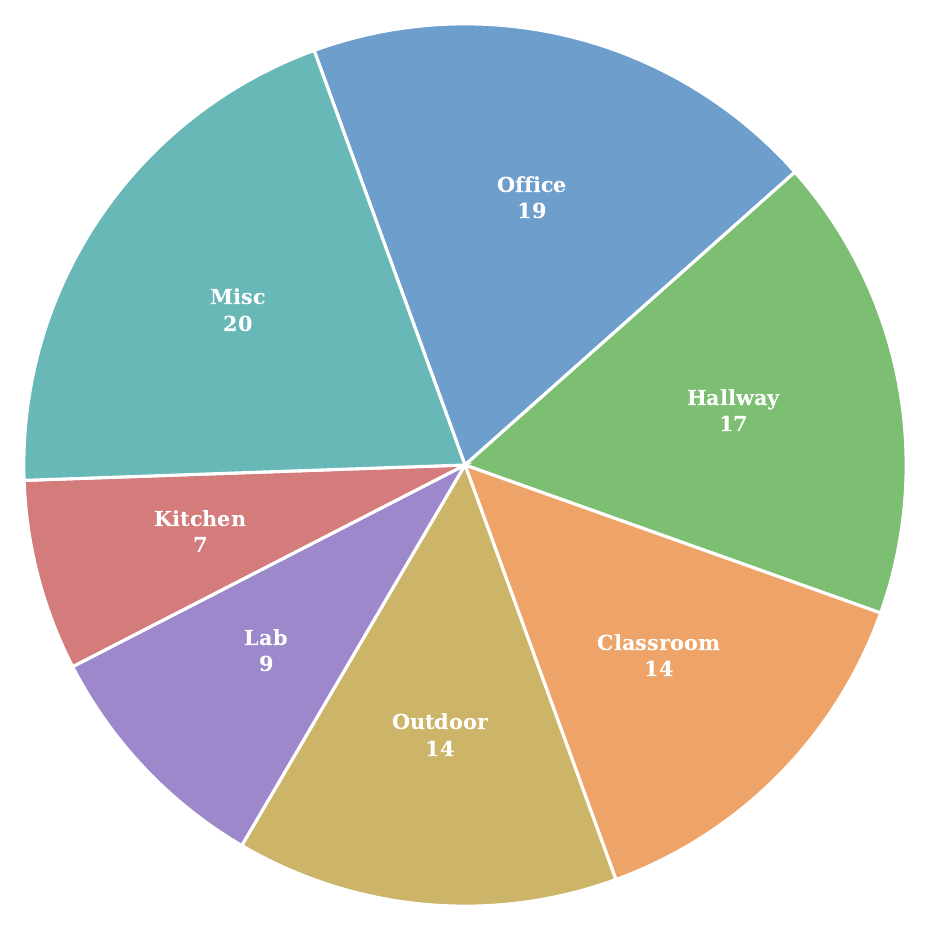}
    \caption{Scene category distribution of our \benchmarkname\ benchmark. \benchmarkname\ covers diverse indoor and outdoor scenes commonly seen in daily life.}
    \label{fig:scene_distribution}
\end{figure}

\begin{figure}[!t]
    \centering
\includegraphics[width=\linewidth]{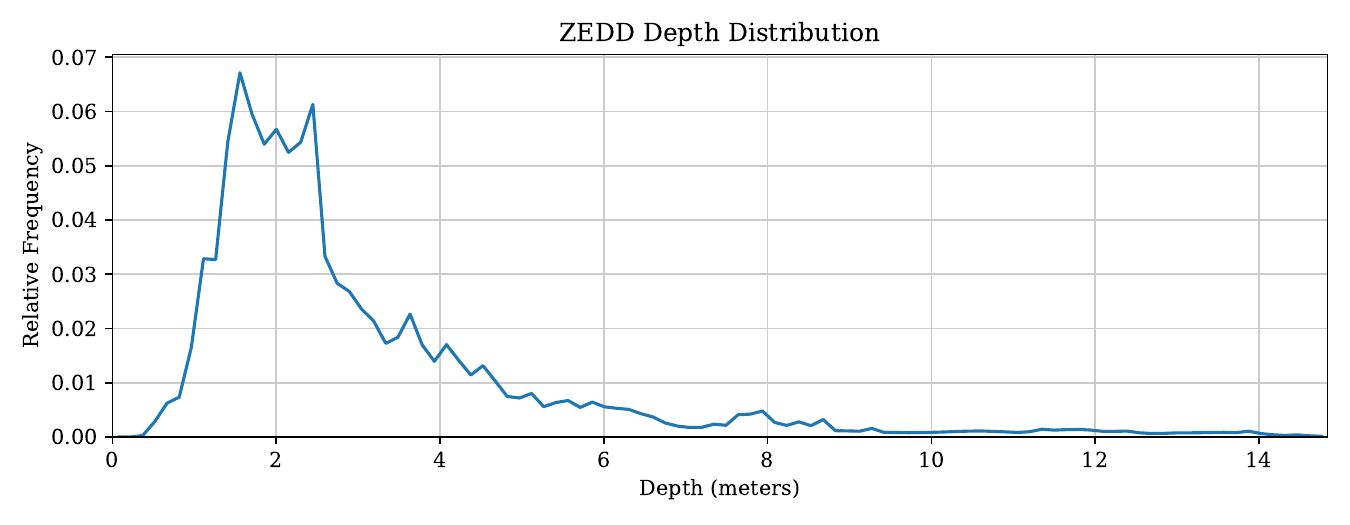}
    \caption{Depth value histogram of the \benchmarkname\ benchmark. \benchmarkname\ covers a diverse depth range from 0.3m to 15m, making it a challenging dataset for depth estimation.}
    \label{fig:depth_histogram}
\end{figure}

\benchmarkname\ is diverse in scene types coverage, as shown in \cref{fig:scene_distribution}. It also covers a wide range of depth values, as shown in \cref{fig:depth_histogram}.

\subsection{Focus Stack Alignment}
\begin{figure}[!b]
    \centering    \includegraphics[width=\linewidth]{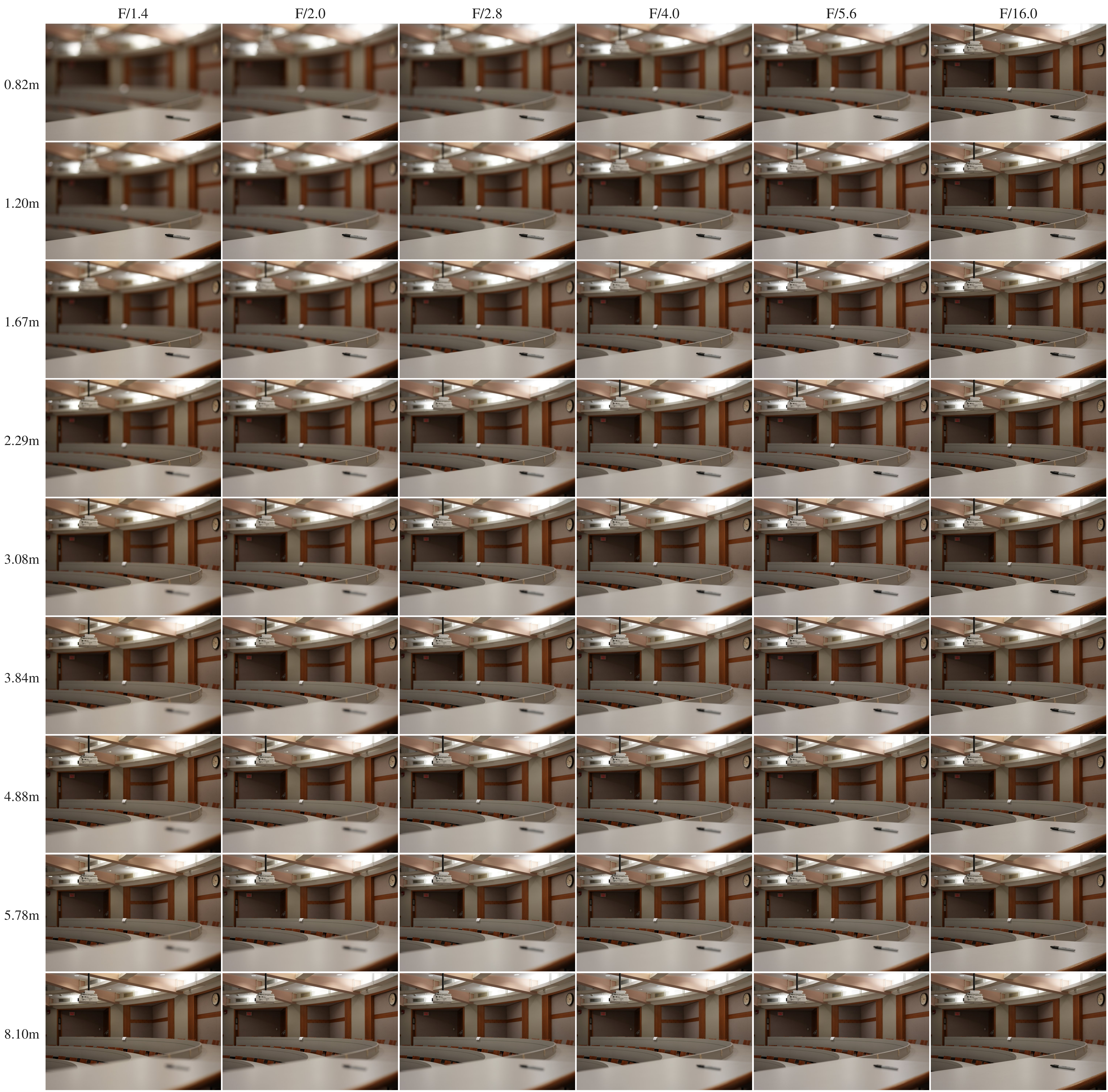}
    \caption{Visualizations of all the 54 images we captured for one scene. Each row is a different focus distance, and each column is a different aperture. The lens breathing is perfectly canceled out by the image-space warping.}
    \label{fig:full_focus_stack}
\end{figure}

We align all the images to a canonical space through image-space warping to compensate for the lens breathing effect. We choose the image captured at focus distance $= 3.08$m as the canonical space (the 5th row in \cref{fig:full_focus_stack}), and we use an open-source image stacking toolbox~\cite{focusstacktoolbox} to perform the alignment. 

We find that directly aligning the large-aperture images yields unstable results, as some images are very blurry, so no reliable image keypoint can be used for matching. Therefore, we first align all the images at the F/16 aperture, and then reuse the image transformation to warp images captured at the same focus distance as the F/16 image but with larger apertures.

A visualization of a full focus stack can be found in \cref{fig:full_focus_stack}. All images are well-aligned. Defocus effects variations can be clearly seen when changing the focus distances and the aperture.

\section{Infinigen Defocus}

\begin{figure}[t]
    \centering
    \includegraphics[width=\linewidth]{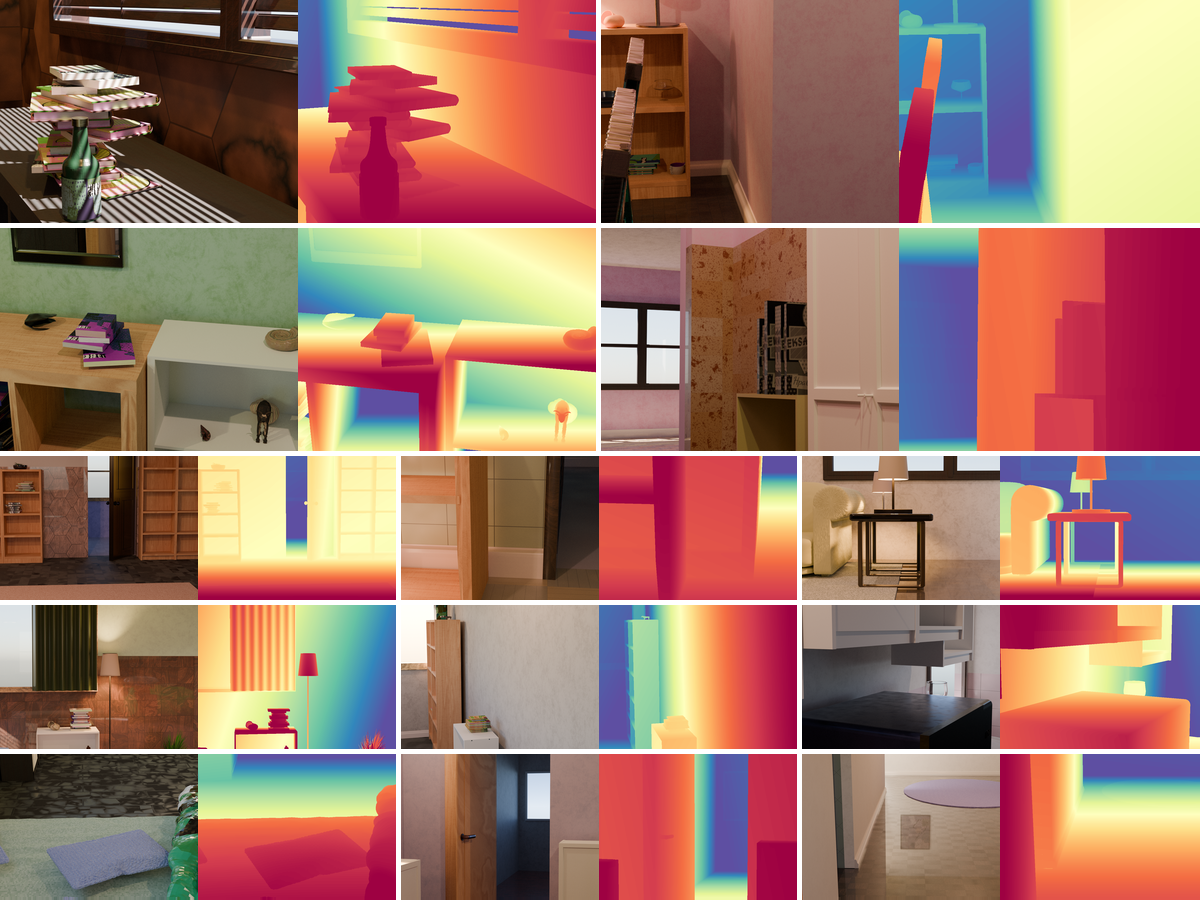}
    \caption{A gallery of the images and depth ground-truth for our Infinigen Defocus Benchmark. The scenes are diverse, and the ground-truth is pixel-perfect.}
    \label{fig:infinigen_gallery}
\end{figure}

We build the Infinigen Defocus synthetic dataset on top of Infinigen Indoors~\cite{infinigen-indoors}. Infinigen is a procedural system for generating photorealistic indoor scenes. Owing to its procedural nature, it can produce unlimited variation at both the object and scene levels, yielding diverse shapes, layouts, and spatial compositions.

Infinigen uses Blender~\cite{blender} for scene composition and rendering. Blender provides native support for camera aperture and focus distance, and supports synthesizing defocus effects during ray tracing using a thin-lens camera model \cite{optics}. This makes Blender suitable for generating realistic focus stacks with physically accurate defocus blur.

We modify the Infinigen generation pipeline so that, for each scene, it renders multiple images from the same camera pose while varying the aperture size and focus distance. We choose the rendering settings to match the distribution covered by \benchmarkname. Specifically, we render images using 5 aperture settings (F1.4/2.0/2.8/4.0/5.6), 9 focus distances (0.8/1.2/1.7/2.3/3.0/3.8/4.7/6.0/8.0m), and one additional all-in-focus image, resulting in $5 \times 9 + 1 = 46$ images per scene. We use the rendered depth map of the all-in-focus image as the ground-truth depth. In total, we generate $500$ scenes and manually reject the scenes with degenerated object layout or suboptimal camera placement, resulting in $200$ scenes with the highest visual quality. A gallery of our Infinigen Defocus dataset are shown in \cref{fig:infinigen_gallery}.

\section{More Qualitative Results \& MobileDepth}
\label{sec:mobiledepth}
\begin{figure}[h]
    \centering
    \includegraphics[width=\linewidth]{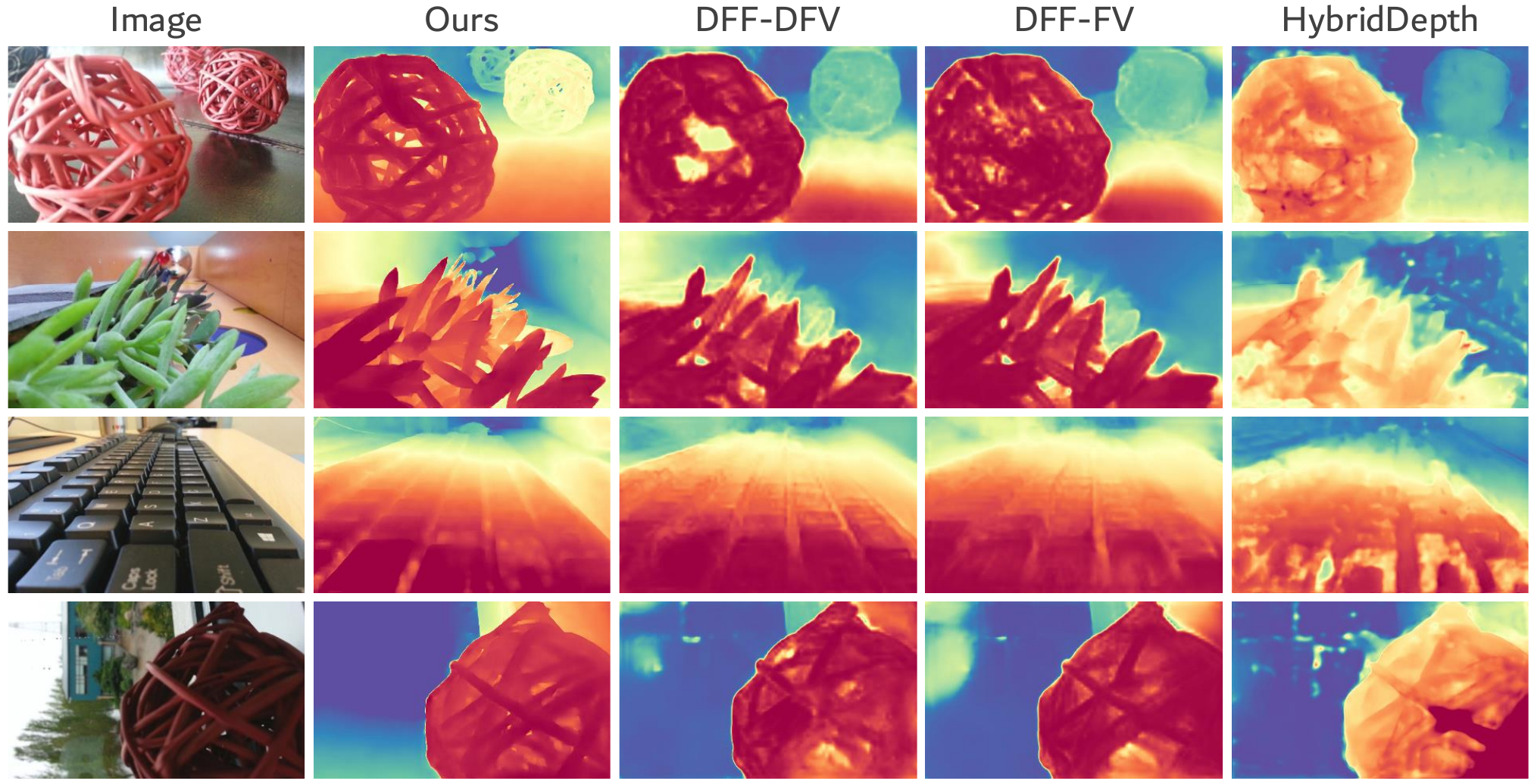}
    \caption{Qualitative comparisons on the MobileDepth~\cite{mobiledepth} dataset.}
    \label{fig:mobiledepth}
\end{figure}

MobileDepth~\cite{mobiledepth} contains 11 scenes with mobile phone captured focus stacks, but no ground-truth depth. Therefore, we only perform a qualitative comparison with DfD baselines. Results are shown in \cref{fig:mobiledepth}. Our method features finer details than all baselines.

\begin{figure}[!b]
    \centering    \includegraphics[width=\linewidth]{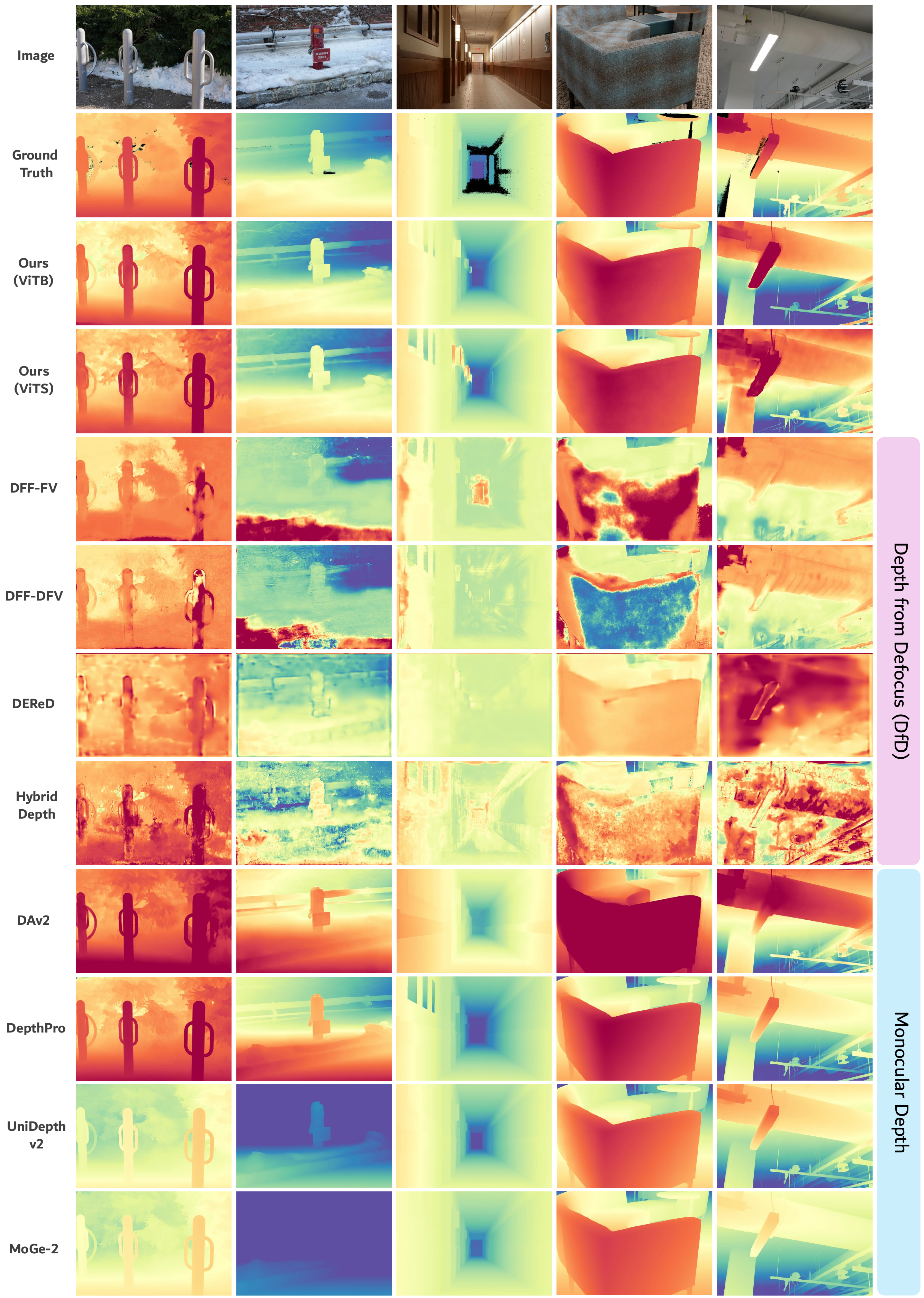}
    \caption{More qualitative comparisons with baselines on \benchmarkname. Takeaways: \textbf{1)} ViT-B produces better details than ViT-S. \textbf{2)} Previous DfD methods fail catastrophically in this zero-shot setting. \textbf{3)} Our method features better metric scale and equally good sharpness as monocular baselines.}
    \label{fig:more_vis}
\end{figure}

We also show more qualitative comparisons with baselines on \benchmarkname\ in \cref{fig:more_vis}. Our method produces sharper details and better metric scale.

\section{Model Training}
\label{sec:appendix_training_details}

\subsection{Training Hyperparameters}
\label{sec:training_hyp}
We use different learning rates for the pretrained backbone parameters and other layers.
Specifically, parameters belonging to the pretrained encoder are scaled by a factor of $0.5$ relative to the base learning rate:
\[
\eta_{\text{pretrained}} = 0.5 \cdot \eta .
\]

The learning rate schedule follows a polynomial decay defined as
\begin{equation}
\eta_t = \eta_0 \left(1 - \frac{t}{T}\right)^{p},
\end{equation}
where $\eta_0 = 5\times10^{-6}$ is the initial learning rate, $t$ is the current training step, $T$ is the total number of training steps, and $p=0.9$ controls the decay rate. We use the AdamW~\cite{loshchilov2017decoupled} optimizer with momentum parameters $\beta_1=0.9$ and $\beta_2=0.999$, and a weight decay of $0.01$. 

We use a consistent sample size of 66K per epoch, regardless of the size of the original datasets. We sample from Hypersim~\cite{hypersim} and TartanAir~\cite{tartanair} with equal probability.

\subsection{Focal Length Randomization}
We train on the image resolution of $518 \times 700$. To get the model familiar with different focal lengths during training, we apply a random zooming augmentation with a zoom factor $s \sim \text{U}(1.0, 1.5)$, where we upscale the image by the factor $s$ and center crop back to  $518 \times 700$. For test images with a different resolution than training, we resize them so that the shorter edge has a length of 518, and resize the depth prediction to the original resolution.

\subsection{Blur Kernel Randomization}
\label{sec:blur_kernel_randomization}

\begin{figure}[t]
\centering
\includegraphics[width=\linewidth]{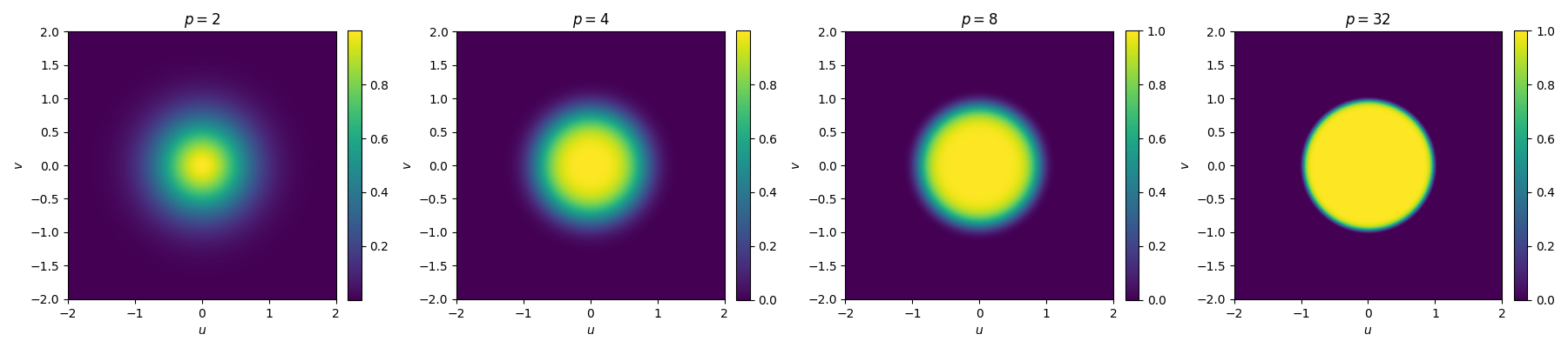}
\caption{Varying the blur kernel shape via the shape parameter $p$ of the generalized PSF. Note that the total energy is not normalized to 1 in this graph.}
\label{fig:disk-blur-varying-on-p}
\end{figure}

To synthesize the focus stack during training, we use a point spread function (PSF) with the circle of confusion (CoC) computed from the camera parameters and the depth map. Given an image $\mathbf{I}$ with ground truth depth $\mathbf{D}$, the CoC $c$ is calculated as:
\begin{equation}
    c = \frac{|\mathbf{D} - d|}{\mathbf{D}} \frac{f^2}{N(d-f)},
\end{equation}
where $d$ is the focus distance, $f$ is the focal length, and $N$ is the f-number. Based on CoC, we compute a point spread function (PSF) as a pixel-wise blur kernel, which is applied to $\mathbf{I}$ through convolution.

There are two common choices for the shape of the PSF: the Gaussian PSF~\cite{dered,gur2019single} and the Disk PSF~\cite{wang2023implicit,talegaonkar2025repurposing}. When the blur radius is small, diffraction dominates and the PSF resembles an Airy pattern~\cite{airy1835diffraction} that is often approximated by a Gaussian. For larger defocus, geometric optics dominates and the PSF approaches a uniform disk corresponding to the circle of confusion. However, neither model alone accurately captures the full range of defocus behavior observed in real images, motivating our generalized PSF formulation.

In the Gaussian PSF, the shape of the blur kernel is a 2D Gaussian with the fallout rate described by $c$: 
\begin{equation}
    \mathcal{F}^{Gaussian} (u,v) = \frac{1}{c^2}\exp \left( -2\frac{u^2+v^2}{c^2} \right),
\end{equation}
whereas the Disk PSF has a uniform energy within the disk: 

\begin{equation}
\mathcal{F}^{Disk}(u,v) =
\begin{cases}
\frac{1}{c^2}, & u^2 + v^2 \le c^2 \\
0, & u^2 + v^2 > c^2,
\end{cases}
\end{equation}
where $u,v$ is a particular pixel offset to the center of the kernel.

We generalize the Gaussian PSF and Disk PSF into a single formula:

\begin{equation}
\mathcal{F}^{Generalized}(u,v,p)
= \frac{1}{c^{2}}
\exp\!\left(
-2\left(\frac{u^{2}+v^{2}}{c^{2}}\right)^{\dfrac{p}{2}}
\right).
\end{equation}

It is easy to prove that the Gaussian PSF and the Disk PSF are special cases when $p=2$ and $p=+\infty$, respectively. The effect on the PSF shape when changing $p$ can be visualized in \cref{fig:disk-blur-varying-on-p}.

We sample $p \sim 2^{\text{U}(1,5)}$ during training. We also randomize the size of the aperture: 
\begin{equation}
    N \sim \text{Random}([1.0, 1.4, 2.0, 2.8, 4.0])
\end{equation}

\subsection{Focus Distance Randomization}
\label{sec:fd_randomization}

To sample an FD list, we need to make two design choices: \textbf{1)} what are the near and far ranges, and \textbf{2)} using what distribution to sample multiple FD within this range.

\noindent\textbf{1) What are the near and far ranges?}
In real-world applications, the valid depth range for focus bracketing can arise from two common scenarios. 
\begin{itemize}
    \item \textbf{Photographer in-the-loop.} The photographer manually specifies a near and far depth range that covers the region of interest in the scene. This allows the focus stack to concentrate sampling density within the depths that matter most for the subject.
    \item \textbf{Fully automatic program.} In automated capture systems, the near and far focus ranges are predetermined and do not depend on the specific scene content. 
\end{itemize}

To cover the first case, we sample the nearest and farthest FD as the 5th and 95th percentiles of the ground-truth depth map for each scene. For the second case, we sample the near and far bounds as:

\begin{equation}
    z_{\text{near}} \sim \text{U}(0.6, 1.0), \;
z_{\text{far}} = z_{\text{near}} \cdot \text{U}(7.0, 15.0).
\end{equation}

We mix these two cases with a ratio of 1:4 during training.

\noindent\textbf{2) Using what distribution to sample FD?} We use a power-law sampling in the disparity space:

\begin{equation}
f_i = \left(
(1 - t_i)\,z_{\text{near}}^{-\kappa}
+ t_i\,z_{\text{far}}^{-\kappa}
\right)^{-1/\kappa},
\;
t_i = \frac{i-1}{S-1}, \quad i = 1,\dots,S,
\end{equation} 
where $S$ is the focus stack size, and $kappa$ is a shape parameter $
\kappa \sim \text{U}(0,1).$

This sampling strategy is intuitive because when $\kappa=1$, the interpolation becomes uniform in $z^{-1}$, \ie, uniform sampling in disparity space, a common strategy adopted in many previous works~\cite{dered,ddff,dfv}. Sampling 
$\kappa \sim \text{U}(0,1)$ further introduces variation in the sampling density, allowing the distribution to smoothly range between nearly uniform-in-log-depth ($\kappa=0$) and uniform-in-disparity ($\kappa=1$), which improves robustness to different focus distance sampling configurations in the real world.

\section{Evaluation Datasets and Baselines}
\label{sec:appendix_datasets_and_baselines}
\noindent\textbf{Real-World RGBD Datasets.} We use the Gaussian PSF (\cref{sec:blur_kernel_randomization}) and the photographer in-the-loop FD (\cref{sec:fd_randomization}) to synthesize the focus stack from the RGBD input on iBims~\cite{ibims}, DIODE~\cite{diode} and HAMMER~\cite{hammer}.

These datasets often have incomplete depth maps, either due to occlusions, low-reflectance surfaces not capturable by depth sensors, or regions too far away. Those missing regions cause trouble when synthesizing the defocus image, as the synthesized image is unsmooth and contains a lot of artifacts if we naively assume a constant depth value in the missing regions. Therefore, we use an off-the-shelf depth completion network OMNI-DC~\cite{omnidc} to fill the missing regions. Note that the filled depth values are only used to synthesize the focus stack but does not affect the evaluation metrics, because we only compute the metrics on the valid depth pixels in ground-truth. 

\noindent\textbf{Baselines} Previous DfD baselines provide multiple checkpoints trained on different datasets. We report the one that works the best empirically, detailed as follows:

\begin{itemize}
    \item HybridDepth~\cite{hybriddepth}: We use the \textit{NYU-PretrainedDFV-FocalStack5} checkpoint.
    \item DFF-DFV and DFF-FV~\cite{dfv}: We use their checkpoints trained on a mixture of FoD500~\cite{defocusnet} and DDFF~\cite{ddff}.
    \item DeRED~\cite{dered}: We use their checkpoint trained on NYUv2~\cite{nyuv2}.
\end{itemize}

Finally, for DFF-DFV finetuned using our training data,  we apply the same loss as ours and only on the finest granularity depth prediction. We also used the same data augmentations, scheduler, and optimizer as ours (\cref{sec:training_hyp}).

\section{Limitations and Future Works}

Firstly, our model is only trained to deal with static scenes. If the camera or objects move during the focus stack capture process, the focus stack won't be perfectly aligned anymore. A future direction is to improve the robustness of DfD on dynamic scenes or handheld cameras, potentially by using optical flow. 

Secondly, the way we synthesize the focus stack using PSF during training creates a sim-to-real domain gap. Although our domain randomization technique on PSF shapes greatly alleviates the problem, there is still space for improvement. One potential direction is to use learning-based image synthesis models such as BokehMe~\cite{bokehme} to create more realistic defocus effects.

\clearpage

\clearpage

\end{document}